\documentclass[10pt,twocolumn,letterpaper]{article}

\usepackage[pagenumbers]{cvpr} 

\usepackage[dvipsnames]{xcolor}

\usepackage{epigraph}
\usepackage{graphicx}
\usepackage{amsmath}
\usepackage{amssymb}
\usepackage{mathtools}
\usepackage{booktabs}
\usepackage{algorithm}
\usepackage{algorithmic}
\usepackage{wrapfig}
\usepackage{glossaries}
\usepackage{colortbl}
\usepackage{bbding}
\usepackage{tikz}
\usepackage{comment}
\usepackage{color}
\usepackage{xspace}
\usepackage{booktabs}
\usepackage{nicefrac}
\usepackage{bbm}
\usepackage{bm}
\usepackage{tabularx,verbatim}
\usepackage{multirow}
\usepackage[accsupp]{axessibility}  
\usepackage{pifont}
\usepackage{tikz}

\definecolor{remark}{rgb}{1,.5,0} 
\definecolor{citecolor}{rgb}{0,0.443,0.737} 
\definecolor{linkcolor}{rgb}{0.956,0.298,0.235} 
\definecolor{cyan}{rgb}{0.831,0.901,0.945}

\usepackage{newfloat}
\usepackage{listings}
\DeclareCaptionStyle{ruled}{labelfont=normalfont,labelsep=colon,strut=off} 
\floatstyle{ruled}
\newfloat{listing}{tb}{lst}{}
\floatname{listing}{Listing}

\usepackage{amsmath,amsfonts,bm}

\def\eqref#1{equation~\ref{#1}}

\def\1{\bm{1}}

\def\rmS{{\mathbf{S}}}

\def\rmU{{\mathbf{U}}}
\def\rmV{{\mathbf{V}}}

\DeclareMathAlphabet{\mathsfit}{\encodingdefault}{\sfdefault}{m}{sl}
\SetMathAlphabet{\mathsfit}{bold}{\encodingdefault}{\sfdefault}{bx}{n}

\DeclareMathOperator*{\argmax}{arg\,max}

\definecolor{me_node}{RGB}{16,119,51}
\definecolor{sup_node}{RGB}{0, 90, 181}
\definecolor{sub_node}{RGB}{212, 17, 89}

\definecolor{higher}{RGB}{16,119,51}
\definecolor{lower}{RGB}{220,50,32}

\definecolor{ours_meth}{RGB}{16,119,51}
\definecolor{bl_meth}{RGB}{115,144,167}

\definecolor{firstBest}{rgb}{0.9, 1, 0.9}
\definecolor{secondBest}{rgb}{1, 0.95, 0.95}

\newcommand{\tablestyle}[2]{\setlength{\tabcolsep}{#1}\renewcommand{\arraystretch}{#2}\centering\footnotesize}

\newcommand{\hierarchy}{H\xspace}

\newcommand{\ovodlong}{open-vocabulary object detection\xspace}
\newcommand{\ovod}{OvOD\xspace}
\newcommand{\shine}{SHiNe\xspace}

\newcommand{\inattitle}{iNaturalist Localization 500\xspace}

\newcommand{\inat}{iNatLoc\xspace}

\newcommand{\fsodtitle}{Few-shot Object Detection dataset\xspace}

\newcommand{\fsod}{FSOD\xspace}

\newcommand{\vlmlong}{vision-language model\xspace}
\newcommand{\vlm}{VLM\xspace}

\newcommand{\llmlong}{large language model\xspace}
\newcommand{\llm}{LLM\xspace}

\newcommand{\chils}{CHiLS\xspace}
\newcommand{\hclip}{H-CLIP\xspace}

\newcommand{\bNexus}{\mathbf{N}}
\newcommand{\bnexus}{\mathbf{n}}

\newcommand{\hier}{\mathcal{H}}
\newcommand{\sent}{e}

\newcommand{\enctxt}{\mathcal{E}_{\text{txt}}}

\newcommand{\bI}{\mathbf{I}}
\newcommand{\bb}{\mathbf{b}}

\newcommand{\bz}{\mathbf{z}}

\newcommand{\bw}{\mathbf{w}}
\newcommand{\bW}{\mathbf{W}}

\newcommand{\classSup}{\mathcal{C}^{\mathrm{det}}}
\newcommand{\dataSup}{\mathcal{D}^{\mathrm{det}}}
\newcommand{\classWeak}{\mathcal{C}^{\mathrm{weak}}}
\newcommand{\dataWeak}{\mathcal{D}^{\mathrm{weak}}}
\newcommand{\classTest}{\mathcal{C}^{\mathrm{test}}}

\newcommand{\vldet}{VLD\xspace}

\definecolor{cvprblue}{rgb}{0.21,0.49,0.74}

\usepackage[pagebackref,breaklinks,colorlinks,citecolor=cvprblue]{hyperref}

\usepackage[capitalize]{cleveref}
\crefname{section}{Sec.}{Secs.}
\Crefname{section}{Section}{Sections}
\Crefname{table}{Table}{Tables}
\crefname{table}{Tab.}{Tabs.}

\def\paperID{1464} 
\def\confName{CVPR}
\def\confYear{2024}

\newcommand{\lblsec}[1]{\label{sec:#1}}
\newcommand{\lblfig}[1]{\label{fig:#1}}
\newcommand{\lbltab}[1]{\label{tbl:#1}}

\newcommand{\lbleq}[1]{\label{eq:#1}}
\newcommand{\lblalg}[1]{\label{alg:#1}}
\newcommand{\refsec}[1]{Sec.~\ref{sec:#1}}
\newcommand{\reffig}[1]{Fig.~\ref{fig:#1}}
\newcommand{\reftab}[1]{Tab.~\ref{tbl:#1}}

\newcommand{\refalg}[1]{Alg.~\ref{alg:#1}}

\newcommand{\refsupp}[1]{App.~\ref{sec:#1}}

\newcommand{\myparagraph}[1]{\vspace{0.06cm}\noindent\textbf{#1}}



\newcommand{\mytexttt}[1]{{\small \texttt{#1}}}

\newcommand{\lesspace}{\vspace{-0.4cm}}

\definecolor{baseline_fig1}{RGB}{194, 46,  90}
\definecolor{ours_fig1}{RGB}{37, 89,  175}

\definecolor{butter1}{RGB}{252, 233,  79}
\definecolor{butter2}{RGB}{237, 212,   0}
\definecolor{butter3}{RGB}{196, 160,   0}
\colorlet{LightButter}{butter1}
\colorlet{Butter}{butter2}
\colorlet{DarkButter}{butter3}

\definecolor{orange1}{RGB}{252, 175,  62}
\definecolor{orange2}{RGB}{245, 121,   0}
\definecolor{orange3}{RGB}{206,  92,   0}
\colorlet{LightOrange}{orange1}
\colorlet{Orange}{orange2}
\colorlet{DarkOrange}{orange3}

\definecolor{chocolate1}{RGB}{233, 185, 110}
\definecolor{chocolate2}{RGB}{193, 125,  17}
\definecolor{chocolate3}{RGB}{143,  89,   2}
\colorlet{LightChocolate}{chocolate1}
\colorlet{Chocolate}{chocolate2}
\colorlet{DarkChocolate}{chocolate3}

\definecolor{chameleon1}{RGB}{138, 226,  52}
\definecolor{chameleon2}{RGB}{115, 210,  22}
\definecolor{chameleon3}{RGB}{ 78, 154,   6}
\colorlet{LightChameleon}{chameleon1}
\colorlet{Chameleon}{chameleon2}
\colorlet{DarkChameleon}{chameleon3}

\definecolor{skyblue1}{RGB}{114, 159, 207}
\definecolor{skyblue2}{RGB}{ 52, 101, 164}
\definecolor{skyblue3}{RGB}{ 32,  74, 135}
\colorlet{LightSkyBlue}{skyblue1}
\colorlet{SkyBlue}{skyblue2}
\colorlet{DarkSkyBlue}{skyblue3}

\definecolor{plum1}{RGB}{173, 127, 168}
\definecolor{plum2}{RGB}{117,  80, 123}
\definecolor{plum3}{RGB}{ 92,  53, 102}
\colorlet{LightPlum}{plum1}
\colorlet{Plum}{plum2}
\colorlet{DarkPlum}{plum3}

\definecolor{scarletred1}{RGB}{239,  41,  41}
\definecolor{scarletred2}{RGB}{204,   0,   0}
\definecolor{scarletred3}{RGB}{164,   0,   0}
\colorlet{LightScarletRed}{scarletred1}
\colorlet{ScarletRed}{scarletred2}
\colorlet{DarkScarletRed}{scarletred3}

\definecolor{aluminium1}{RGB}{238, 238, 236}
\definecolor{aluminium2}{RGB}{211, 215, 207}
\definecolor{aluminium3}{RGB}{186, 189, 182}
\definecolor{aluminium4}{RGB}{136, 138, 133}
\definecolor{aluminium5}{RGB}{ 85,  87,  83}
\definecolor{aluminium6}{RGB}{ 46,  52,  54}

\definecolor{indigo}{RGB}{114,  33, 188}
\definecolor{maroon}{RGB}{103,   7,  72}
\definecolor{turquoise}{RGB}{ 64, 224, 208}
\definecolor{green4}{RGB}{  0, 139,   0}

\definecolor{DarkCoral}{rgb}{0.8, 0.36, 0.27}

\crefname{section}{\S}{\S\S}
\crefname{subsection}{\S}{\S\S}
\crefformat{table}{Table~#2#1#3}
\crefformat{figure}{Fig.~#2#1#3}
\crefformat{equation}{Eq.~#2#1#3}
\newcommand{\rowNumber}[1]{}
\definecolor{highlightRowColor}{rgb}{0.95, 0.95, 1}
\definecolor{baselineRowColor}{rgb}{0.95, 0.95, 0.95}
\definecolor{firstBest}{rgb}{0.9, 1, 0.9}
\definecolor{secondBest}{rgb}{1, 0.95, 0.95}

\definecolor{correctpred}{RGB}{0,153,77}
\definecolor{makesensepred}{RGB}{209, 70, 153}
\definecolor{wrongpred}{RGB}{250, 0, 0}
\definecolor{superpred}{RGB}{0, 127, 255}

\def\eg{\emph{e.g.\,}}

\def\ie{\emph{i.e.\,}}

\def\etal{\emph{et al.\,}}

\title{SHiNe: Semantic Hierarchy Nexus for Open-vocabulary Object Detection}

\author{Mingxuan Liu$^{1,2}$\thanks{Correspondence to: {\tt\small mingxuan.liu@unitn.it}~~~~~~~~}~~~~
Tyler L. Hayes$^{2}$~~~~
Elisa Ricci$^{1,3}$~~~~
Gabriela Csurka$^{2}$~~~~
Riccardo Volpi$^{2}$\\
$^{1}$ University of Trento \quad\quad\quad $^{2}$ NAVER LABS Europe \quad\quad\quad $^{3}$ Fondazione Bruno Kessler\\
}

\begin{document}
\maketitle

\begin{abstract}
Open-vocabulary object detection (\ovod) has transformed detection into a language-guided task, empowering users to freely define their class vocabularies of interest during inference. However, our initial investigation indicates that existing \ovod detectors exhibit significant variability when dealing with vocabularies across various semantic granularities, posing a concern for real-world deployment. To this end, we introduce \textbf{S}emantic \textbf{Hi}erarchy \textit{\textbf{Ne}xus} (\textbf{\shine}), a novel classifier that 
uses
semantic knowledge from class hierarchies. It 
runs
\textit{offline} in three steps: \textit{i)} it retrieves relevant super-/sub-categories from a hierarchy for each target class; \textit{ii)} it integrates these categories into hierarchy-aware sentences; \textit{iii)} it fuses these sentence embeddings to generate the \textit{nexus} classifier vector. Our evaluation on various detection benchmarks demonstrates that \shine enhances robustness across diverse vocabulary granularities, achieving up to +31.9\% mAP50 with 
ground truth
hierarchies, while retaining improvements using hierarchies generated by large language models. Moreover, when applied to open-vocabulary classification on ImageNet-1k, \shine improves the CLIP zero-shot baseline by +2.8\% accuracy. \shine is training-free and can be seamlessly integrated with any \textit{off-the-shelf} \ovod detector, without incurring 
additional computational overhead
during inference. 
The code is  \href{https://github.com/naver/shine}{open source}.

\end{abstract}    
\vspace{-8.0mm}
\epigraph{\textit{A complicated series of connections between different things.}}{\small {Definition of \textit{Nexus}}, \textbf{Oxford Dictionary}}
\vspace{-8.0mm}

\section{Introduction}
\lblsec{introduction}

\begin{figure}[ht!]
    \centering
    \includegraphics[width=0.9\linewidth]{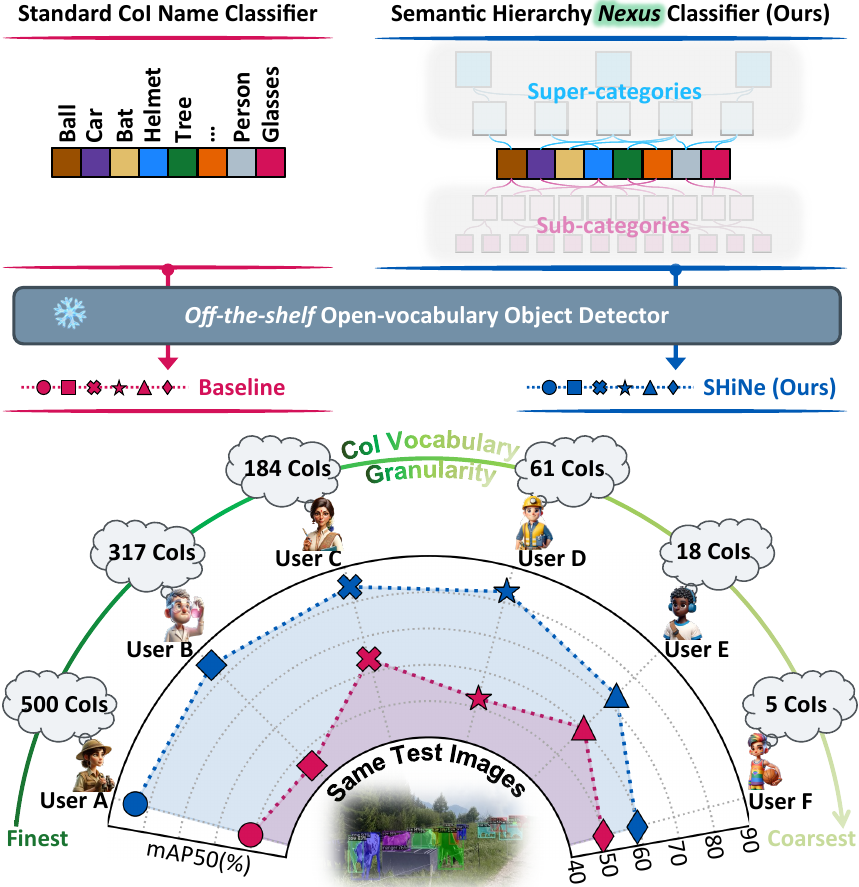}
    \vspace{+2.0mm}
    \caption{
    \textbf{(Top)} Classifier comparison for open-vocabulary object detectors: \textbf{(Left)} standard methods use solely class names in the vocabulary specified by the user to extract text embeddings; \textbf{(Right)} our proposed \shine fuses information from super-/sub-categories into \textit{nexus} points to generate hierarchy-aware representations. \textbf{(Bottom)} Open-vocabulary detection performance at different levels of vocabulary granularity specified by users: A standard {\color{baseline_fig1}Baseline} under-performs and presents significant variability; {\color{ours_fig1} SHiNe} allows for improved and more uniform performance across various vocabularies. Results are on the \inat~\cite{cole2022label} dataset.
    }
    \lblfig{teaser}
\vspace{-4.0mm}
\end{figure}

Open-vocabulary object detection (\ovod)~\cite{gu2021open, zareian2021open, zhu2023survey,WuPAMI23TowardsOpenVocabularyLearning} transforms the object detection task into a language-guided matching problem between visual regions and class names. 
Leveraging 
weak supervisory signals and a pre-aligned vision-language space from Vision-Language Models (\vlm{s})~\cite{radford2021learning,jia2021scaling}, \ovod methods~\cite{zhu2023survey, zhou2022detecting, gu2021open, zareian2021open, lin2022learning} extend the ability {of models} to localize and categorize objects beyond the trained categories. 
Under the \ovod paradigm, target object classes are described using text prompts like \mytexttt{"a \{Class Name\}"}, rather than class indices. By altering the \mytexttt{"\{Class Name\}"}, \ovod methods enable users to \textit{freely} define their own Classes of Interest (CoIs) using natural language. This allows new classes of interest to be detected without the need for model re-training.

Yet, recent studies for open-vocabulary classification~\cite{novack2023chils,parashar2023prompting,ge2023improving} highlight a key challenge: open-vocabulary methods are sensitive to the choice of vocabulary. For instance, Parashar \etal~\cite{parashar2023prompting} enhanced CLIP's zero-shot performance by substituting scientific CoI names, like \mytexttt{"Rosa"}, with common English names, such as \mytexttt{"Rose"}. Recent \ovod models have improved performance by better aligning object features with the \vlm semantic space~\cite{wu2023aligning, kuo2022f}. However, a pivotal question remains: \textit{Are off-the-shelf \ovod detectors truly capable of handling an open vocabulary \textbf{across various semantic granularities}?}

In practical scenarios, Classes of Interest (CoIs) are in the eyes of the beholder. For example, consider a region crop of a \mytexttt{"Dog"}: one user may be interested in the specific breed (\textit{e.g.}, \mytexttt{"Labrador"}), while another might only be concerned about whether it is an \mytexttt{"Animal"}. Thus, the CoI is defined at varying levels of semantic granularity. Ideally, since these CoIs refer to the same visual region, the performance of an \ovod detector should be consistent across different granularities. However, our initial experiments (illustrated in \reffig{teaser}) reveal that the performance of an \ovod detector~\cite{zhou2022detecting} (see {\color{baseline_fig1} \textbf{Baseline}}) fluctuates based on the vocabulary granularity. 
This inconsistency in performance across granularities presents a significant concern for deploying \textit{off-the-shelf} \ovod models in real-world contexts, especially in safety-critical~\cite{knight2002safety} areas like autonomous driving~\cite{martinez2018autonomous}.

Although the same physical object, a \mytexttt{"Labrador"}, can be classified at varying levels of granularity, the inherent \textit{fact} that a \mytexttt{"Labrador is a dog, which is an animal"} remains \textit{constant}. This knowledge is readily available from a semantic hierarchy. Guided by this rationale, we aim to enhance the robustness of existing \ovod detectors to vocabularies specified at any granularity by leveraging knowledge inherent in semantic hierarchies. Recent research in open-vocabulary classification~\cite{novack2023chils,ge2023improving} has explored using super-/sub-categories of CoIs from hierarchies to improve accuracy. However, these methods involve searching through sub-categories or both super-/sub-categories at inference time, leading to 
additional
computational overhead and limiting their use in detection tasks.

We introduce the \textbf{S}emantic \textbf{Hi}erarchy \textit{\textbf{Ne}xus} (\textbf{\shine}), a novel classifier designed to enhance the robustness of \ovod to diverse vocabulary granularities. \shine is \textit{training-free}, and ensures that the inference procedure is \textit{linear} in complexity relative to the number of CoIs.
\shine first retrieves relevant super(abstract)-/sub(specific)-categories from a semantic hierarchy for each CoI in a vocabulary. It then uses an \textbf{Is-A} connector to integrate these categories into hierarchy-aware sentences, while \textit{explicitly} modeling their internal relationships. 
Lastly, it fuses these text embeddings into a vector, termed \textit{nexus}, using an aggregator (\textit{e.g.},
the mean operation)
to form a classifier weight for the target CoI. 
\shine can be directly integrated with any \textit{off-the-shelf} \vlm-based \ovod detector. As shown in \reffig{teaser}, \shine consistently improves performance across a range of CoI vocabulary granularities, while narrowing performance gaps at different granularities.

We evaluate \shine on various detection datasets~\cite{cole2022label, fan2020few}, that cover a broad range of label vocabulary granularities.
This includes scenarios with readily available hierarchies and cases \textit{without} them. In the latter, we utilize \llmlong{s}~\cite{chatgpt} to generate a synthetic~\cite{novack2023chils} three-level hierarchy for \shine. Our results demonstrate that \shine significantly and consistently improves the performance and robustness of baseline detectors, and showcase its generalizability to other \textit{off-the-shelf} \ovod detectors. Additionally, we extend \shine to open-vocabulary classification and further validate its effectiveness by comparing it with two state-of-the-art methods~\cite{novack2023chils, ge2023improving} on the ImageNet-1k~\cite{deng2009imagenet} dataset. The key contributions of this work are:
\begin{itemize}
    \item We show that the performance of existing \ovod detectors varies across vocabulary granularities. This highlights the need for enhanced robustness to 
    arbitrary
    granularities, especially for real-world applications.
    
    \item We introduce \shine, a novel classifier that improves the robustness of \ovod models to various vocabulary granularities using semantic knowledge from hierarchies. \shine is \textit{training-free} and compatible with existing and generated hierarchies. It can be seamlessly integrated into any \ovod detector \textit{without} 
    computational overhead.
    
    \item We demonstrate that \shine consistently enhances the performance 
    of \ovod detectors across various vocabulary granularities on \inat~\cite{cole2022label} and \fsod~\cite{fan2020few}, with gains of up to {\color{higher}\textbf{+31.9}} points
    in mAP50. 
    On open-vocabulary classification, \shine improves the CLIP~\cite{radford2021learning} zero-shot baseline by up to {\color{higher}\textbf{+2.8\%}} 
    on ImageNet-1k~\cite{deng2009imagenet}.
\end{itemize}
\section{Related Work}
\lblsec{relatedwork}
\myparagraph{Open-vocabulary object detection (OvOD)}~\cite{zhu2023survey,WuPAMI23TowardsOpenVocabularyLearning} is rapidly gaining traction due to its practical significance, allowing users to \textit{freely} define their Classes of Interest (CoIs) during inference and facilitating the detection of newly specified objects in a zero-shot way. With the aid of weak supervisory signals, OvOD surpasses zero-shot detectors~\cite{tan2021survey} by efficiently aligning visual region features with an embedding space that has been \textit{pre-aligned} with image and text by contrastive \vlmlong{s} (\vlm{s})~\cite{radford2021learning, jia2021scaling}. This process is approached from either the vision or text side to bridge the gap between region-class and image-class alignments. To this end, methods based on region-aware training~\cite{zareian2021open,zang2022open,yao2022detclip,wu2023cora}, pseudo-labeling~\cite{zhou2022detecting,zhong2022regionclip,feng2022promptdet,arandjelovic2023three}, knowledge distillation~\cite{gu2021open,du2022learning,wu2023aligning}, and transfer learning~\cite{kuo2022f,minderer2023scaling,zhang2023simple} are explored. In our study, we 
apply our method to pre-trained region-text aligned \ovod detectors,
improving their performance and robustness to vocabularies of diverse granularities. 
Our method shares conceptual similarities with the work of Kaul~\etal~\cite{kaul2023multi}, where they develop a multi-modal classifier that merges a text-based classifier enriched with descriptors~\cite{menon2022visual} from GPT-3~\cite{brown2020language} and a vision classifier grounded in image exemplars. This classifier is then used to train an \ovod detector \cite{zhou2022detecting} with an extra \textit{learnable} bias. In contrast, our proposed \shine is \textit{training-free}, enabling effortless integration with any OvOD detector.

\myparagraph{Prompt engineering}~\cite{gu2023systematic} has been extensively studied as a technique to enhance \vlm{s}~\cite{radford2021learning,jia2021scaling,zhang2023multi}. \textit{Prompt enrichment} methods~\cite{menon2022visual, pratt2023does, parashar2023prompting, roth2023waffling, yan2023learning} have focused on augmenting frozen \vlm text classifiers by incorporating additional class descriptions sourced from \llmlong{s} (\llm{s})~\citep{brown2020language}. In contrast, our work explores the acquisition of useful semantic knowledge from a hierarchy. \textit{Prompt tuning} methods~\cite{zhou2022learning,shu2022test,khattak2023maple,zhou2022conditional,ren2023prompt, wang2023transhp} introduced \textit{learnable} token vectors into text prompts, which are fine-tuned on downstream tasks. In contrast, our proposed method is \textit{training-free}. Our work is mostly related to two recent methods, CHiLS~\cite{novack2023chils} and H-CLIP~\cite{ge2023improving}, that improve CLIP's~\cite{radford2021learning} zero-shot classification performance by relying on a semantic hierarchy. CHiLS searches for higher logit score matches within the sub-categories, using the max score found to update the initial prediction. H-CLIP runs a combinatorial search over related super-/sub-categories prompt combinations for higher logit scores. However, both approaches incur 
additional
computational overhead due to their \textit{search-on-the-fly} mechanism during inference, constraining their use to classification tasks. In contrast, \shine operates offline and adds no
overhead at inference, 
{making it}
applicable to both classification and detection tasks.

\myparagraph{Semantic hierarchy}~\cite{fellbaum1998wordnet,van2018inaturalist,cole2022label,wah2011caltech} is a tree-like taxonomy~\cite{wu2005learning} or a directed acyclic graph~\cite{ruiz2002hierarchical} that structures semantic concepts following an \textit{asymmetric} and \textit{transitive} ``Is-A'' relation~\cite{silla2011survey}. Previous works have 
used
such hierarchies to benefit various vision tasks~\cite{barz2019hierarchy,deng2010does,frome2013devise,goodman2001classes,morin2005hierarchical,ruggiero2015higher,bertinetto2020making}. Cole \etal~\cite{cole2022label} introduce the extensive iNatLoc dataset with a six-level hierarchy to enhance weakly supervised object localization, showing that appropriate label granularity can improve model training. Shin 
\etal~\cite{shin2020hierarchical} and Hamamci \etal~\cite{hamamci2023diffusion} develop hierarchical architectures that incorporate multiple levels of a label hierarchy for training, enhancing multi-label object detection in remote sensing and dental X-ray images, respectively. Our work distinguishes itself from previous studies in {two} key 
ways: \textit{i)} We focus on multi-modal models; \textit{ii)} We improve 
OvOD detectors using label hierarchies as an external knowledge base, without
requiring
hierarchical annotations or
any training. Furthermore, \shine does not rely on a ground-truth hierarchy and can work with an LLM-generated~\cite{chatgpt} hierarchy.
\section{Method}
\lblsec{mthd_method}

\begin{figure*}[!t]
    \centering
    \includegraphics[width=\linewidth]{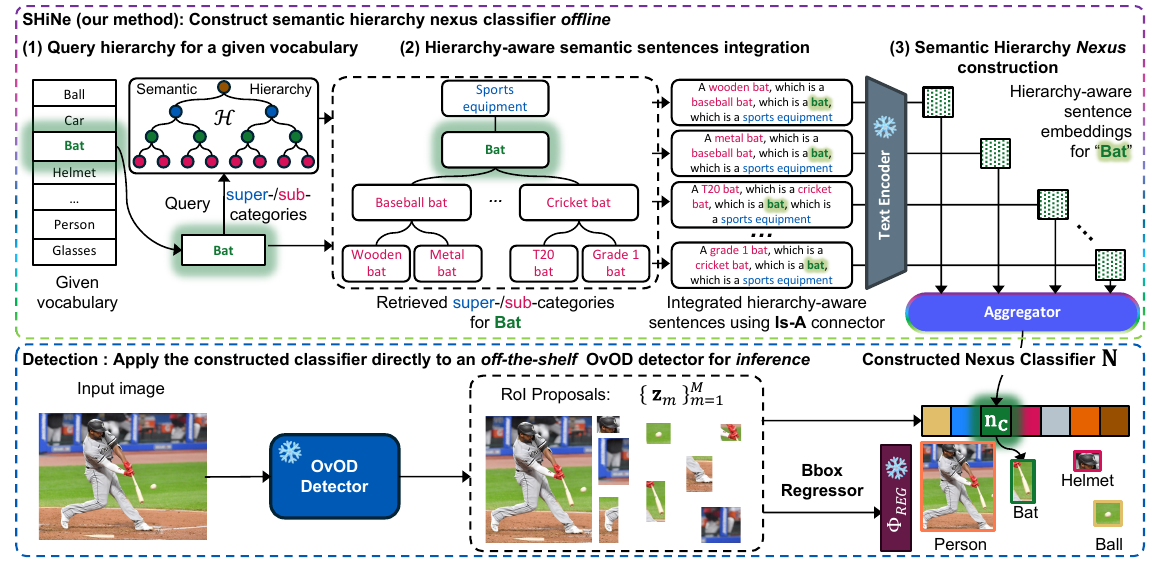}
    \vspace{-7.0mm}
    \caption{
    Overview of our method. \textbf{(Top)} \shine constructs the semantic hierarchy \textit{nexus} classifier in three steps \textit{offline}: (1) For each target class (\textit{e.g.}, \mytexttt{"{\color{me_node}Bat}"} in {\color{me_node}green}) in the given vocabulary, we query the associated super-(in \text{{\color{sup_node}blue}})/sub-(in \text{{\color{sub_node}pink}}) categories from a semantic hierarchy. (2) These retrieved categories along with their interrelationships are integrated into a set of hierarchy-aware sentences using our proposed \textbf{Is-A} connector. (3) These sentences are then encoded by a frozen VLM text encoder (\textit{e.g.}, CLIP~\cite{radford2021learning}) and subsequently fused using an aggregator (\textit{e.g.}, mean-aggregator) to form a \textit{nexus} classifier vector for the target class. \textbf{(Bottom)}: The constructed classifier is directly applied to an \textit{off-the-shelf} \ovod detector for inference, enhancing its robustness across various levels of vocabulary granularity.
    }
    \lblfig{approach}
    \lesspace
\end{figure*}

Our objective is to improve the robustness of \textit{off-the-shelf} open-vocabulary object detectors to diverse user-defined Classes of Interest (CoIs) with varying levels of semantic granularity. We first provide an introduction of open-vocabulary object detection (\ovod). \refsec{mthd_shine} introduces our method of developing the \textbf{S}emantic \textbf{Hi}erarchy \textit{\textbf{Ne}xus} (\shine) based classifier for \ovod detectors to improve their vocabulary granularity robustness. Once established, the \shine classifier can be directly integrated with existing trained \ovod detectors and transferred to novel datasets in a zero-shot manner as discussed in \refsec{mthd_zeroshot}.

\lblsec{mthd_preliminary}
\myparagraph{Problem formulation.}
The objective of \ovodlong is to localize and classify novel object classes freely specified by the user within an image, without 
any
retraining, in a zero-shot manner. Given an input image $\bI \in \mathbb{R}^{3\times h \times w}$, \ovod localizes all foreground objects and classifies them by estimating a set of bounding box coordinates and class label pairs $\{\bb_m, c_m\}_{m=1}^{M}$, 
with $\bb_m \in \mathbb{R}^4$ and $c_m \in \classTest$, where $\classTest$ is the vocabulary set defined by the user at test time.
To attain open-vocabulary capabilities, \ovod~\cite{zhu2023survey, zhou2022detecting,lin2023match} uses a box-labeled dataset $\dataSup$ with a limited vocabulary $\classSup$ and an auxiliary dataset $\dataWeak$ as weak supervisory signals. $\dataWeak$ features fewer detailed image-class or image-caption annotation pairs, yet it encompasses a broad vocabulary $\classWeak$ (\textit{e.g.}, ImageNet-21k~\cite{deng2009imagenet}), significantly expanding the detection lexicon.

\myparagraph{Open-vocabulary detector.}
Predominant \ovod detectors, such as Detic~\cite{zhou2022detecting} and VLDet~\cite{lin2023match}, follow a two-stage 
pipeline.
First,
given an image, a learned region proposal network (RPN) yields a bag of $M$ region proposals by $\{\bz_m\}_{m=1}^{M} = \Phi_\textsc{RPN}(\bI)$, where $\bz_m \in \mathbb{R}^D$ is a $D$-dimensional region-of-interest (RoI) feature embedding. Then,
for each proposed region, a learned bounding box regressor predicts the location coordinates by $\hat{\bb}_m = \Phi_{\textsc{REG}}(\bz_m)$.
An open-vocabulary classifier estimates a set of classification scores $s_{m}(c,\bz_m) = \langle \bw_{c}, \bz_m \rangle$ for each class, where $\bw_{c}$ is a vector in the classifier $\bW \in \mathbb{R}^{|\classTest|}$ and $\langle \cdot, \cdot \rangle$ is the cosine similarity function. $\bW$ is the frozen text classifier, created by using a \vlm text encoder (\textit{e.g.}, CLIP~\cite{radford2021learning}) to encode the names of CoIs in $\classTest$ specified by the user. The CoI that yields the highest score is assigned as the classification result. During training, \ovod detectors learn all model parameters except for the frozen text classifier. This allows them to achieve region-class alignment by leveraging the vision-language semantic space pre-aligned by \vlm{s} for the open-vocabulary capability. 
Our work aims to improve existing
pre-trained
\ovod detectors, so we omit further details, {and refer the reader to dedicated surveys}~\cite{zhu2023survey,WuPAMI23TowardsOpenVocabularyLearning}.

\subsection{\shine: Semantic Hierarchy Nexus}
\lblsec{mthd_shine}
Here, we 
describe
\shine, 
our proposed
semantic hierarchy \textit{nexus}-based classifier for improving \ovod. As illustrated in \reffig{approach}(top), for each target CoI $c \in \classTest$ (\textit{e.g.}, \mytexttt{"{\color{me_node}Bat}"}) in the user-defined vocabulary, we construct a \textit{nexus} point $\bnexus_c \in \mathbb{R}^{D}$ by {incorporating}
information from related super-/sub-categories derived from a semantic hierarchy $\hier$. \shine is \textit{training-free}. Upon constructing the \textit{nexus} points for the entire vocabulary \textit{offline}, the \textit{nexus}-based classifier $\bNexus$ is directly applied to an \textit{off-the-shelf} \ovod detector for inference. This replaces the conventional CoI name-based classifier $\bW$ with our hierarchy-aware \shine classifier. This enables the classification score $s_{m}(c,\bz_m) = \langle \bnexus_{c}, \bz_m \rangle$ to be high when the proposed region closely aligns with the semantic hierarchy ``theme'' embodied by the \textit{nexus} point. {This point}
represents the fusion of a set of hierarchy-aware semantic sentences from specific to abstract that are relevant to the CoI $c$. Next, we detail the construction process.

\myparagraph{Querying the semantic hierarchy.}
To obtain related super-/sub-categories, a semantic hierarchy $\hier$ is crucial for our approach. In this study, we investigate two types of hierarchies: \textit{i)} dataset-specific class taxonomies~\cite{fan2020few, deng2009imagenet, cole2022label}, and \textit{ii)} hierarchies synthesized for the target test vocabulary using \llmlong{s} (\llm). To generate the synthetic hierarchy, we follow Novack \etal~\cite{novack2023chils} and query an LLM such as ChatGPT~\cite{chatgpt} to generate super-categories ($p=3$) and sub-categories ($q=10$) for each CoI $c \in \classTest$, creating a three-level hierarchy $\hier$
(see \refsupp{supp_hier} for details).
With the hierarchy available, as depicted in \reffig{approach}(1), for each target CoI $c$, we retrieve \textit{all} the related super-/sub-categories, which can assist in distinguishing $c$ from other concepts in the vocabulary across granularities~\cite{ge2023improving}. Note that we exclude the root node (\textit{e.g.}, \mytexttt{"entity"}) from this process, as it does not help differentiate $c$ from other categories.

\myparagraph{Hierarchy-aware semantic sentence integration.}
The collected categories contain {both abstract and specific}
semantics useful for guiding the classification process. However, methods like simple ensembling~\cite{novack2023chils} or concatenation~\cite{ge2023improving} overlook some valuable knowledge \textit{implicitly} provided by the hierarchy, namely the inherent internal relationships among concepts. Inspired by the hierarchy structure definition~\cite{silla2011survey}, we propose an \textbf{Is-A} connector to \textit{explicitly} model these \textbf{interrelationships}. Specifically, for each target CoI $c$, the \textbf{Is-A} connector 
{integrates the retrieved categories into sentences}
from the lowest sub-category (more specific) to the highest super-category (more abstract), including the target CoI name. As depicted in \reffig{approach}(2), this process yields a set of $K$ hierarchy-aware sentences $\{\sent_{k}^{c}\}_{k=1}^{K}$. Each sentence $\sent_{k}^{c}$ contains knowledge that spans from specific to abstract, all related to the target CoI and capturing their inherent relationships, as

\noindent \mytexttt{A {\color{sub_node}wooden baseball bat}, which \textbf{is a} {\color{sub_node}baseball bat}, which \textbf{is a} {\color{me_node}bat}, which \textbf{is a} {\color{sup_node}sports equipment}.}

where the sub-categories, target category, and super-categories are color-coded in {\color{sub_node} red}, {\color{me_node} green}, and {\color{sup_node} blue}.

\myparagraph{Semantic hierarchy \textit{Nexus} construction.}
A \textit{nexus} $\bnexus_c \in \mathbb{R}^{D}$ serves as a unifying embedding that fuses the hierarchy-aware knowledge contained in the integrated sentences $\{\sent_{k}^{c}\}_{k=1}^{K}$. As shown in \reffig{approach}(3), we employ a frozen \vlm~\cite{radford2021learning} text encoder $\enctxt$ to translate the integrated sentences into the region-language aligned semantic space compatible with the downstream \ovod detector. The semantic hierarchy \textit{nexus} for the CoI $c$ is then constructed by aggregating these individual sentence embeddings as:
\begin{align}
    \bnexus_c &= 
    \text{Aggregator}
    \left(
        \left\{
        \enctxt\left(\sent_k^c\right)
        \right\}_{k=1}^{K}
    \right) \enspace
    \lbleq{aggregation}
\end{align}
where, by default, we employ a straightforward but effective \textbf{mean-aggregator} to compute the mean vector of the set of sentence embeddings. The goal of the aggregation process is to fuse the expressive and granularity-robust knowledge into the \textit{nexus} vector, as a ``theme'', from the encoded hierarchy-aware sentences. Inspired by text classification techniques in Natural Language Processing (NLP)~\cite{li1998classification, shin2018interpreting, gewers2021principal}, we also introduce an alternative aggregator, where we perform SVD decomposition of the sentence embeddings and replace the mean vector with the principal eigenvector as $\bnexus_c$. We study its effectiveness in \refsec{expt_selfstudy} and provide a detailed description in \refsupp{supp_impl_peigen}.

\subsection{Zero-shot Transfer with \shine}
\lblsec{mthd_zeroshot}
As shown in \reffig{approach}(bottom), once the \textit{nexus} points are constructed for each CoI in the target vocabulary, the \shine classifier $\bNexus$ can be directly applied 
to the \ovod detector
for inference, assigning class names to proposed regions as:
\vspace{-4.0mm}
\begin{align}
    \hat{c}_m = \argmax_{c \in \classTest} \langle \bnexus_{c}, \bz_m \rangle
    \lbleq{classifier_nexus} \enspace
\end{align}
where $\bz_m$ is the $m$-th region embedding.
Given that $\bnexus_c \in \mathbb{R}^{D}$, it becomes evident from Eq.~\ref{eq:classifier_nexus} that \shine {has the same}
computational complexity as the vanilla name-based \ovod classifier.
Let us note that \shine is not limited to detection, it can be adapted to open-vocabulary classification by
substituting the region embedding $\bz_m$ with an image one. 
We validate this claim by also benchmarking on ImageNet-1k~\cite{deng2009imagenet}.
We provide the pseudo-code and time complexity analysis of \shine in \refsupp{supp_impl_code} and \refsupp{supp_impl_time}, {respectively.}
\section{Experiments}
\lblsec{ovod_expt}
\begin{table}[!h]
    \centering
    \footnotesize
    \tablestyle{2.4pt}{0.8}
    \caption{
    Evaluation dataset descriptions of \inat and \fsod. Label granularity ranges from finest (F) to coarsest (C) levels.
    }
    \begin{tabular}{c|ccl|ccl}
         \toprule
         \multirow{2}{*}{\rotatebox[origin=c]{90}{Gran}}
         & \multicolumn{3}{c|}{\inat} 
         & \multicolumn{3}{c}{\fsod}\\
         

         & Level & \# Classes & \multicolumn{1}{c|}{Label Example} 
         & Level & \# Classes & \multicolumn{1}{c}{Label Example}\\

         \midrule

         \multirow{6}{*}{\rotatebox[origin=c]{90}{C $\xleftarrow{\hspace{0.8cm}}$ F}}
         & 6 & 500 & Cyprinus carpio 
         & \multirow{2}{*}{3} & \multirow{2}{*}{200} & \multirow{2}{*}{Watermelon} \\
         
         & 5 & 317 & Cyprinus 
         &  &  & \\
         
         & 4 & 184 & Cyprinidae 
         & \multirow{2}{*}{2} & \multirow{2}{*}{46} & \multirow{2}{*}{Fruit} \\
         
         & 3 & 64  & Cypriniformes 
         &  &  & \\
         
         & 2 & 18  & Actinopterygii 
         & \multirow{2}{*}{1} & \multirow{2}{*}{15} & \multirow{2}{*}{Food} \\
         
         & 1 & 5   & Chordata 
         &  &  & \\
         \bottomrule
    \end{tabular}
    \lesspace
    \lbltab{inat_fsod_hier}
\end{table}
\begin{table}[!h]
    \centering
    \footnotesize
    \tablestyle{8pt}{0.8}
    \caption{
    Training signal combinations. LVIS~\cite{gupta2019lvis} and COCO~\cite{lin2014microsoft} are used as strong box-level supervision. ImageNet-21k~\cite{deng2009imagenet} (IN-21k) and the 997-class subset (IN-L) of ImageNet-21k that overlaps with LVIS are used as weak image-level supervision.
    }
    \begin{tabular}{c|cc}
         \toprule
          Notation & Strong Supervision & Weak Supervision\\
         \midrule
         \textbf{I}   & LVIS & N/A           \\
         \textbf{II}  & LVIS & IN-L          \\
         \textbf{III} & LVIS & IN-21k        \\
         \textbf{IV}  & LVIS \& COCO & IN-21k \\
         \bottomrule
    \end{tabular}
    \lbltab{training_data}
    \lesspace
\end{table}

\myparagraph{Evaluation protocol and datasets.} We primarily follow the cross-dataset transfer evaluation (CDTE) protocol~\cite{zhu2023survey} in our experiments. In this scenario, the \ovod detector is trained on one dataset and then tested on other datasets in a zero-shot manner. This enables a thorough evaluation of model performance across diverse levels of 
vocabulary granularity. We conduct experiments on two detection datasets: \inattitle (\inat)~\cite{cole2022label} and \fsodtitle (\fsod)~\cite{fan2020few}, which have ground-truth hierarchies for evaluating object labeling at multiple levels of granularity.
\inat is a fine-grained detection dataset featuring a consistent six-level label hierarchy based on the 
{biological}
tree of life, along with bounding box annotations for its validation set. \fsod is assembled from OpenImages~\cite{kuznetsova2020open} and ImageNet~\cite{deng2009imagenet}, structured with a two-level label hierarchy. For a more comprehensive evaluation, we use \fsod's test split and manually construct one more hierarchy level atop its existing top level, resulting in a three-level label granularity for evaluation. \reftab{inat_fsod_hier} outlines the number of label hierarchy levels and the corresponding category counts for both datasets, accompanied by examples to demonstrate the semantic granularity. Detailed dataset statistics and their hierarchies are available in \refsupp{supp_data}. We use the mean Average Precision (mAP) at an Intersection-over-Union (IoU) threshold of 0.5 (mAP50) as our main evaluation metric. Additional experiments on COCO~\cite{lin2014microsoft} and LVIS~\cite{gupta2019lvis} under the open-vocabulary protocol~\cite{gu2021open} are provided in \refsupp{supp_expt_cocolvis}.

\begin{figure*}[!th]
    \centering
    \includegraphics[width=\linewidth]{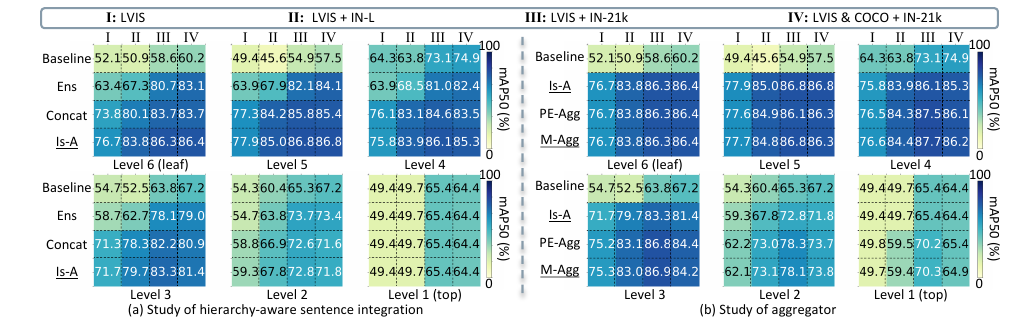}
    \vspace{-7.0mm}
    \caption{
    Study of hierarchy-aware sentence integration methods \textbf{(left)} and aggregators \textbf{(right)} across various label granularity levels on the \inat dataset. Detic with a Swin-B backbone is used as the baseline. Darker background color indicates higher mAP50. The default components of \shine are \underline{underlined}. {Note that the experiment in (a) omits sub-categories and the aggregation step.}
    }
    \vspace{-6.0mm}
    \lblfig{self_study}
\end{figure*}

\myparagraph{Baseline detector.}
In our experiments, we use the pre-trained Detic~\cite{zhou2022detecting} method as the baseline detector, given its open-source code and strong performance. Detic is a two-stage \ovod detector that relies on CenterNet2~\cite{zhou2021probabilistic} and incorporates a frozen text classifier generated from the CLIP ViT-B/32 text encoder~\cite{radford2021learning} using a prompt of the form: \mytexttt{"a \{Class\}"}. Detic uses both detection and classification data (image-class weak supervisory signals) for training. In our experiments, we explore and compare with Detic under four variants of supervisory signal combinations as shown in \reftab{training_data}. We study a ResNet-50~\cite{he2016deep} and a Swin-B~\cite{liu2021swin} backbone pre-trained on ImageNet-21k-P~\cite{ridnik2021imagenet}.

\myparagraph{\shine implementation details.}
To directly apply our method to the baseline \ovod detector, we use the CLIP ViT-B/32~\cite{radford2021learning} text encoder to construct the \shine classifier and directly apply it to the baseline \ovod detector, following the pipeline described in \refsec{mthd_shine}. We use the mean-aggregator by default. In our experiments, we employ and study two sources for the hierarchy: the ground-truth hierarchy structure provided by the dataset and a synthetic hierarchy generated by an \llm. We use the
gpt-3.5-turbo model~\cite{chatgpt} as our \llm via its public API to produce a simple 3-level hierarchy (comprising one child and one parent level) for the given target CoI vocabulary with temperature 0.7, as outlined in \refsec{mthd_method}. We detail the hierarchy generation process in \refsupp{supp_hier} and report the statistics.

\begin{table*}[!t]
    \centering
    \small
    \setlength{\tabcolsep}{2.0pt} 
    \caption{
    Detection performance across varying label granularity levels, ranging from finest (F) to coarsest (C), on \inat \textbf{(upper)} and \fsod \textbf{(lower)} datasets. \shine is directly applied to the baseline detector (BL)~\cite{zhou2022detecting} with ground-truth (GT-\hierarchy) and LLM-generated (LLM-\hierarchy) hierarchies. ResNet-50~\cite{he2016deep} \textbf{(left)} and Swin-B~\cite{liu2021swin} \textbf{(right)} backbones~\cite{liu2021swin} are compared. Four types of supervisory signal combinations are investigated. Note (\dag): At the L1-/L6-level of GT-\hierarchy, no super-/sub-categories categories are used, respectively. mAP50 (\%) is reported.
    }
    \begin{tabular}{lll|lll|lll|lll|lll|}
         \toprule
         &&\multicolumn{1}{c}{}
         & \multicolumn{6}{c}{ResNet-50 Backbone} 
         & \multicolumn{6}{c}{Swin-B Backbone} \\

         \cmidrule(r){4-9}
         \cmidrule(r){10-15}

         &&\multicolumn{1}{c}{}
         & \multicolumn{3}{c}{\textbf{I} - LVIS} 
         & \multicolumn{3}{c}{\textbf{II} - LVIS + IN-L} 
         & \multicolumn{3}{c}{\textbf{III} - LVIS + IN-21k} 
         & \multicolumn{3}{c}{\textbf{IV} - LVIS \& COCO + IN-21k}\\

         \cmidrule(r){4-6}
         \cmidrule(r){7-9}
         \cmidrule(r){10-12}
         \cmidrule(r){13-15}

         \multirow{2}{*}{\rotatebox[origin=c]{90}{Set}} & \multirow{2}{*}{\rotatebox[origin=c]{90}{Gran}} & \multirow{2}{*}{\rotatebox[origin=c]{90}{Level}} 
         & \multicolumn{1}{c}{\multirow{2}{*}{BL}} & \multicolumn{1}{c}{\shine} & \multicolumn{1}{c|}{\shine} 
         & \multicolumn{1}{c}{\multirow{2}{*}{BL}} & \multicolumn{1}{c}{\shine} & \multicolumn{1}{c|}{\shine} 
         & \multicolumn{1}{c}{\multirow{2}{*}{BL}} & \multicolumn{1}{c}{\shine} & \multicolumn{1}{c|}{\shine} 
         & \multicolumn{1}{c}{\multirow{2}{*}{BL}} & \multicolumn{1}{c}{\shine} & \multicolumn{1}{c|}{\shine} \\
         
         &&
         && \multicolumn{1}{c}{(GT-\hierarchy)} & \multicolumn{1}{c|}{(LLM-\hierarchy)} &  & \multicolumn{1}{c}{(GT-\hierarchy)} & \multicolumn{1}{c|}{(LLM-\hierarchy)} & & \multicolumn{1}{c}{(GT-\hierarchy)} & \multicolumn{1}{c|}{(LLM-\hierarchy)} &  & \multicolumn{1}{c}{(GT-\hierarchy)} & \multicolumn{1}{c|}{(LLM-\hierarchy)} \\

         \cmidrule(r){4-6}
         \cmidrule(r){7-9}
         \cmidrule(r){10-12}
         \cmidrule(r){13-15}
         
         \multirow{6}{*}{\rotatebox[origin=c]{90}{\inat}} & \multirow{6}{*}{\rotatebox[origin=c]{90}{(C $\xleftarrow{\hspace{1.2cm}}$ F)}}

         & L6\dag
         & 32.0 & 48.4({\color{higher}+16.4}) & \bf52.8({\color{higher}+20.8}) 
         & 35.2 & 57.1({\color{higher}+21.9}) & \bf58.3({\color{higher}+23.1})
         & 58.6 & \bf86.3({\color{higher}+27.7}) & 84.5({\color{higher}+25.9}) 
         & 60.2 & \bf86.4({\color{higher}+26.2}) & 82.7({\color{higher}+22.5})\\
								
         && L5 
         & 28.2 & \bf49.4({\color{higher}+21.2}) & 41.1({\color{higher}+12.9}) 
         & 30.3 & \bf59.0({\color{higher}+28.7}) & 46.6({\color{higher}+16.3})
         & 54.9 & \bf86.8({\color{higher}+31.9}) & 76.3({\color{higher}+21.4})  
         & 57.5 & \bf86.3({\color{higher}+28.8}) & 76.1({\color{higher}+18.6})\\
 										
         && L4 
         & 40.1 & \bf51.5({\color{higher}+11.4}) & 50.4({\color{higher}+10.3}) 
         & 43.4 & \bf61.4({\color{higher}+18.0}) & 57.5({\color{higher}+14.1})
         & 73.1 & \bf87.7({\color{higher}+14.6}) & 84.0({\color{higher}+10.9})  
         & 74.9 & \bf86.2({\color{higher}+11.3}) & 83.4({\color{higher}+8.5})\\
								
         && L3 
         & 38.8 & 56.5({\color{higher}+17.7}) & \bf57.2({\color{higher}+18.4}) 
         & 41.6 & \bf65.3({\color{higher}+23.7}) & 61.7({\color{higher}+20.1})
         & 63.8 & \bf86.9({\color{higher}+23.1}) & 83.6({\color{higher}+19.8})  
         & 67.2 & \bf84.3({\color{higher}+17.1}) & 81.7({\color{higher}+14.5})\\
				
         && L2 
         & 34.4 & \bf45.0({\color{higher}+10.6}) & 43.9({\color{higher}+9.5}) 
         & 39.3 & \bf53.7({\color{higher}+14.4}) & 50.5({\color{higher}+11.2})
         & 65.3 & \bf78.1({\color{higher}+12.8}) & 77.2({\color{higher}+11.9}) 
         & 67.2 & 73.8({\color{higher}+6.6}) & \bf74.5({\color{higher}+7.3})\\
 											
         && L1\dag 
         & 31.6 & \bf33.6({\color{higher}+2.0}) & 33.5({\color{higher}+1.9}) 
         & 32.5 & \bf43.3({\color{higher}+10.8}) & 36.9({\color{higher}+4.4})
         & 65.4 & \bf70.3({\color{higher}+4.9}) & 63.8({\color{lower}-1.6}) 
         & 64.4 & \bf64.9({\color{higher}+0.5}) & 62.1({\color{lower}-2.3})\\
         
         \cmidrule(r){4-6}
         \cmidrule(r){7-9}
         \cmidrule(r){10-12}
         \cmidrule(r){13-15}
         
         \multirow{3}{*}{\rotatebox[origin=c]{90}{\fsod}} & \multirow{3}{*}{\rotatebox[origin=c]{90}{(C $\leftarrow$ F)}}
 										
         & L3\dag 
         & 49.7 & 52.1({\color{higher}+2.4}) & \bf52.2({\color{higher}+2.5}) 
         & 51.9 & 53.6({\color{higher}+1.7}) & \bf53.7({\color{higher}+1.8})
         & 66.0 & \bf66.7({\color{higher}+0.7}) & 66.3({\color{higher}+0.3}) 
         & 65.6 & \bf66.4({\color{higher}+0.8}) & \bf66.4({\color{higher}+0.8}) \\

         && L2 
         & 28.2 & \bf39.9({\color{higher}+11.7}) & 30.9({\color{higher}+2.7}) 
         & 27.8 & \bf39.8({\color{higher}+12.0}) & 29.8({\color{higher}+2.0})
         & 38.4 & \bf51.4({\color{higher}+13.0}) & 40.3({\color{higher}+1.9}) 
         & 39.4 & \bf52.4({\color{higher}+13.0}) & 41.5({\color{higher}+2.1})\\
         
         && L1\dag 
         & 16.0 & \bf34.3({\color{higher}+18.3}) & 22.0({\color{higher}+6.0}) 
         & 16.5 & \bf31.4({\color{higher}+14.9}) & 21.0({\color{higher}+4.5})
         & 24.7 & \bf42.2({\color{higher}+17.5}) & 30.2({\color{higher}+5.5}) 
         & 25.0 & \bf42.5({\color{higher}+17.5}) & 29.6({\color{higher}+4.6})\\
         
         \bottomrule
    \end{tabular}
    \lesspace
    \lbltab{imprv_detic_swin}
\end{table*}

\subsection{Analysis of \shine}
\lblsec{expt_selfstudy}
We first study the core components of \shine on the \inat~\cite{cole2022label} using its \textit{ground-truth} hierarchy. Consistent findings on the FSOD~\cite{fan2020few} dataset are reported in \refsupp{supp_expt_ablation_fsod}.

\myparagraph{The Is-A connector effectively integrates hierarchy knowledge in natural sentences.}
To assess the effectiveness of our \textbf{Is-A} connector, we design control experiments for constructing the \ovod classifier with a \textit{single} sentence, omitting sub-categories and the aggregation step. Specifically, for a target CoI like \mytexttt{"{\color{me_node}Baseball bat}"}, we retrieve only its super-categories at each ascending hierarchy level. We then explore three ways to integrate the CoI with its ascending super-categories in natural language and create the classifier vector as follows: 
\begin{itemize}
    \item \textbf{Ensemble (Ens)}:
    \{\mytexttt{"{\color{me_node}baseball bat}"}, \mytexttt{"{\color{sup_node}bat}"}, \mytexttt{"{\color{sup_node}sports equipment}"}\}

    \item \textbf{Concatenate (Concat)}:
    \mytexttt{"A {\color{me_node}baseball bat} {\color{sup_node}bat {\color{sup_node}sports equipment}}"}

    \item \textbf{Is-A (Ours)}:
    \mytexttt{"A {\color{me_node}baseball bat}, which \textbf{is a} {\color{sup_node}bat}, which \textbf{is a} {\color{sup_node}sports equipment}"}
\end{itemize}
where the super-categories are colored in {\color{sup_node}blue}. For \textbf{Concat} and \textbf{Is-A}, we create the classifier vector for the target CoI by encoding the \textit{single} sentence with the CLIP text encoder. For the \textbf{Ens} method, we use the average embedding of the ensembled names as: $\frac{1}{3}(\enctxt(\mytexttt{"{\color{me_node}baseball bat}"})+\enctxt(\mytexttt{"{\color{sup_node}bat}"})+\enctxt(\mytexttt{"{\color{sup_node}sports equipment}"}))$. Next, we conduct control experiments to evaluate the three integration methods as well as the standard CoI name-based baseline methods. As shown in \reffig{self_study}(a), except for the top levels where all methods degrade to the standard baseline (no super-category nodes), all methods outperform the baseline across all granularity levels by directing the model's focus towards more abstract concepts via the included super-categories.
Among the methods compared, our \textbf{Is-A} connector excels across all granularity levels, boosting the baseline mAP50 by up to \textbf{\color{higher}+39.4} points (see last row and second column in \reffig{self_study}(a-L5)). This underscores the effectiveness of our \textbf{Is-A} connector, which integrates related semantic concepts into sentences and explicitly models their relationships, yielding hierarchy-aware embeddings.

\myparagraph{A simple mean-aggregator is sufficient for semantic branch fusion.}
We explored two aggregation methods: mean-aggregator (\textbf{M-Agg}) and principal eigenvector aggregator (\textbf{PE-Agg}). Note that in this experiment, all methods use the proposed \textbf{Is-A} connector to create \textit{a set of} hierarchy-aware sentences to aggregate, ranging from each retrieved sub-category to the super-categories, as elaborated in \refsec{mthd_method}. As \reffig{self_study}(b) shows, both methods improve performance over the baseline across various models and label granularities. Note that these aggregators revert to the simple \textbf{Is-A} method at the leaf level where no sub-categories are available for aggregation. The benefits of aggregation methods are more pronounced with coarser granularity, significantly outperforming the baseline and the \textbf{Is-A} method, with gains up to \textbf{\color{higher}+9.8} on \inat (see third row and second column in \reffig{self_study}(b-L1). Notably, \textbf{M-Agg} generally outperforms \textbf{PE-Agg} despite its simplicity, making it the default choice for \shine in the subsequent experiments. Nonetheless, we aim to highlight the effectiveness of \textbf{PE-Agg}: to the best of our knowledge, this is the first study using the principal eigenvector as a classifier vector in vision-language models.

\subsection{\shine on Open-vocabulary Detection}
\lblsec{expt_ovod}

\begin{table}[!t]
    \centering
    \small
    \tablestyle{1.9pt}{0.8}
    \caption{
    Comparison with CoDet~\cite{ma2024codet} and VLDet (\vldet)~\cite{lin2022learning} 
    on \inat and \fsod.
    \shine is applied to the baseline methods, respectively.
    All methods employ Swin-B~\cite{liu2021swin} as backbone. Box-annotated LVIS~\cite{gupta2019lvis} and image-caption-annotated CC3M~\cite{sharma2018conceptual} are used as supervisory signals. mAP50 (\%) is reported.
    }
    \begin{tabular}{ll|lll|lll|}
         \toprule

         \multirow{2}{*}{\rotatebox[origin=c]{90}{Set}} 
         & \multirow{2}{*}{\rotatebox[origin=c]{90}{Level}} 
         & \multicolumn{1}{c}{\multirow{2}{*}{CoDet}} & \multicolumn{1}{c}{\shine} & \multicolumn{1}{c|}{\shine} 
         & \multicolumn{1}{c}{\multirow{2}{*}{\vldet}} & \multicolumn{1}{c}{\shine} & \multicolumn{1}{c|}{\shine} \\
         
         &
         && \multicolumn{1}{c}{(GT-\hierarchy)} & \multicolumn{1}{c|}{(LLM-\hierarchy)}
         &&\multicolumn{1}{c}{(GT-\hierarchy)} & \multicolumn{1}{c|}{(LLM-\hierarchy)}\\

         \cmidrule(r){3-5}
         \cmidrule(r){6-8}
         
         \multirow{6}{*}{\rotatebox[origin=c]{90}{\inat}}

         & L6
         & 48.7 & \bf80.1({\color{higher}+31.4}) & 75.1({\color{higher}+26.4})
         & 81.7 & \bf84.0({\color{higher}+2.3})  & 83.8({\color{higher}+2.1})\\
								
         & L5 
         & 43.2 & \bf80.9({\color{higher}+37.7}) & 63.1({\color{higher}+19.9})
         & 83.7 & \bf84.7({\color{higher}+1.0})  & 82.1({\color{lower}-1.6})\\
 										
         & L4 
         & 64.0 & \bf80.5({\color{higher}+16.5}) & 73.8({\color{higher}+9.8})
         & 82.1 & 84.5({\color{higher}+2.4}) & \bf85.8({\color{higher}+3.7})\\

         & L3 
         & 56.1 & \bf79.3({\color{higher}+23.2}) & 76.7({\color{higher}+20.6})
         & 77.7 & \bf83.9({\color{higher}+6.2})  & 83.3({\color{higher}+5.6})\\
				
         & L2 
         & 61.3 & 65.3({\color{higher}+4.0}) & \bf66.0({\color{higher}+4.7})
         & 71.2 & 75.2({\color{higher}+4.0}) & \bf77.2({\color{higher}+6.0})\\
 											
         & L1
         & 52.3 & \bf54.9({\color{higher}+2.6}) & 50.4({\color{lower}-1.9})
         & 66.1 & 66.7({\color{higher}+0.6})    & \bf71.2({\color{higher}+5.1})\\
         
         \cmidrule(r){3-5}
         \cmidrule(r){6-8}
         
         \multirow{3}{*}{\rotatebox[origin=c]{90}{\fsod}}
         
         & L3
         & 60.5 & \bf62.5({\color{higher}+2.0}) & 61.6({\color{higher}+1.1})
         & 60.5 & \bf63.7({\color{higher}+3.2}) & 63.3({\color{higher}+2.8})\\
 									
         & L2 
         & 33.5 & \bf48.5({\color{higher}+15.0}) & 36.6({\color{higher}+3.1})
         & 33.9 & \bf49.2({\color{higher}+15.3}) & 37.4({\color{higher}+3.5})\\
         
         & L1
         & 19.9 & \bf39.7({\color{higher}+19.8}) & 25.4({\color{higher}+5.5})
         & 20.8 & \bf41.6({\color{higher}+20.8}) & 26.2({\color{higher}+5.4})\\
         
         \bottomrule
    \end{tabular}
    \lesspace
    \lbltab{imprv_vldet_codet_swinb}
\end{table}

\myparagraph{\shine operates with different hierarchies.}
In this section, we broaden our investigation to assess the effectiveness and the robustness of \shine with different semantic hierarchy sources. \reftab{imprv_detic_swin} shows the comparative analysis across various levels of label granularity between the baseline \ovod detector and our method, using either the ground-truth hierarchy or the LLM-generated hierarchy as proxies. We observe that our approach consistently surpasses the baseline by a large margin across all granularity levels on both datasets---and this holds true whether we employ the ground-truth or LLM-generated hierarchy. Averaged across all models and granularity levels on \inat, our method 
yields
an improvement of \textbf{\color{higher}+16.8} points using the ground-truth hierarchy and \textbf{\color{higher}+13.4} points with the LLM-generated hierarchy. For the \fsod dataset, we 
observe
gains of \textbf{\color{higher}+10.3} and \textbf{\color{higher}+2.9} points, respectively. Although the performance gains are smaller with the LLM-generated hierarchy, they nonetheless signify a clear enhancement over the baseline across label granularities on all examined datasets. 
This shows
that \shine is not reliant on ground-truth hierarchies. Even when applied to noisy, synthetic hierarchies, it yields substantial performance improvements. 
{Additional results are in ~\refsupp{supp_expt_more_swinb} and ~\refsupp{supp_expt_stat}.}

\myparagraph{\shine operates with other \ovod detectors.}
To evaluate \shine's generalizability, we apply \shine to additional \ovod detectors: CoDet~\cite{ma2024codet} and VLDet (\vldet)~\cite{lin2022learning}.
The evaluation results showcased in \reftab{imprv_vldet_codet_swinb} affirm that \shine consistently improves the performance of CoDet and VLDet significantly across different granularities on both datasets, with both hierarchies. Further, we assess \shine on another DETR-style~\cite{carion2020end} detector, CORA~\cite{wu2023cora}, in \refsupp{supp_expt_oth_ovod}.
\begin{figure}[!th]
    \vspace{-3.6mm}
    \centering
    \includegraphics[width=\linewidth]{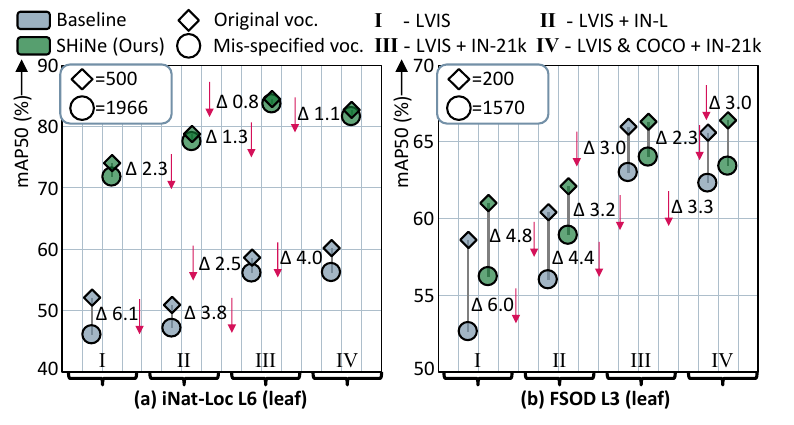}
    \vspace{-6.0mm}
    \caption{
    Analysis of \ovod detection performance under noisy \textit{mis-specified} label vocabularies on \inat \textbf{(left)} and \fsod \textbf{(right)} datasets. We assess the detection performance of both the baseline detector ({\color{bl_meth}in grey}) and our method ({\color{ours_meth}in green}) under varied supervision signals, contrasting results between the original (\rotatebox[origin=c]{45}{$\square$}) and the expanded mis-specified ($\bigcirc$) vocabularies. \shine employs the LLM-generated hierarchy for both vocabularies. We report mAP50, highlighting the performance drop ($\Delta$).
    }
    \lesspace
    \lblfig{expanded_comp}
\end{figure}

\myparagraph{\shine is resilient to \textit{mis-specified} vocabularies.}
\lblsec{expt_ovod_expanded}
In real-world applications, an authentic open vocabulary text classifier may be constructed using a vocabulary comprising a wide array of CoIs, even though only a subset of those specified classes appear in the test data. We define these as \textit{mis-specified} vocabularies. Studying resilience in this challenging scenario is essential for practical applications. To this end, we gathered 500 class names from OpenImages~\cite{kuznetsova2020open} and 1203 from LVIS~\cite{gupta2019lvis}, resulting in 1466 unique classes after deduplication. These are added as ``noisy'' CoIs to the \inat and \fsod \textit{leaf} label vocabularies, creating expanded sets with 1966 and 1570 CoIs, respectively. Using ChatGPT, \shine generates simple 3-level hierarchies for each class in these expanded vocabularies. As shown in \reffig{expanded_comp}, mis-specified vocabularies cause a decrease in baseline detector performance, dropping an average of \textbf{\color{lower}-4.1} points on \inat and \textbf{\color{lower}-4.2} points on \fsod. However, interestingly, \shine not only continues to offer performance gains over the baseline detector but also mitigates the performance drop to \textbf{\color{lower}-1.4} on \inat and \textbf{\color{lower}-3.3} on \fsod, respectively. This suggests that \shine not only improves the robustness but also enhances the resilience of the baseline detector when confronted with a mis-specified vocabulary.

\subsection{\shine on Open-vocabulary Classification}
\lblsec{expt_cls}
In this section, we adapt \shine to open-vocabulary classification, by simply substituting the region embedding in Eq.~\ref{eq:classifier_nexus} with an image embedding from the CLIP image encoder~\cite{radford2021learning}. 
We
evaluate it on the zero-shot transfer classification task 
using
the well-established ImageNet-1k benchmark~\cite{deng2009imagenet}. We compare \shine with two state-of-the-art hierarchy-based methods: \chils~\cite{novack2023chils} and \hclip~\cite{ge2023improving}, which are specifically designed for 
classification.

\begin{table}
    \centering
    \small
    \tablestyle{3.5pt}{0.95}
    \caption{
    ImageNet-1k~\cite{deng2009imagenet} zero-shot classification. We compare with two state-of-the-art hierarchy-based methods under WordNet (WDN) and LLM-generated hierarchies. Vanilla CLIP~\cite{radford2021learning} 
    serves as the baseline. We report top-1 accuracy, and FPS measured on the same NVIDIA RTX 2070 GPU with a batch size 1 and averaged over 10 runs. \dag: For fair comparison, we reproduce H-CLIP's results without its uncertainty estimation step and its \textit{refined} WordNet hierarchy. In the original H-CLIP paper, a top-1 accuracy of 67.78\% on ImageNet-1k was achieved using ViT-B/16 encoders.
    }
    \begin{tabular}{ll|ll|ll|ll}
         \toprule
         \multicolumn{2}{l|}{}
         & \multicolumn{2}{c|}{ViT-B/32} & \multicolumn{2}{c|}{ViT-B/16} & \multicolumn{2}{c}{ViT-L/14} \\

         \cline{3-4}
         \cline{5-6}
         \cline{7-8}

         &
         & \multicolumn{1}{c}{Acc(\%)} & \multicolumn{1}{c|}{FPS} 
         & \multicolumn{1}{c}{Acc(\%)} & \multicolumn{1}{c|}{FPS} 
         & \multicolumn{1}{c}{Acc(\%)} & \multicolumn{1}{c}{FPS} \\
        
         \hline
         \rowcolor{baselineRowColor} &  CLIP
         & 58.9 & 150
         & 63.9 & 152
         & 72.0 & 81 \\
         
         \hline
         
         \multirow{3}{*}{\rotatebox[origin=c]{90}{WDN}}
         
         & H-CLIP\dag
         & 58.7({\color{lower}-0.2}) & 3
         & 63.8({\color{lower}-0.1}) & 3 
         & 70.6({\color{lower}-1.4}) & 2 \\
         
         & CHiLS
         & 59.6({\color{higher}+0.7}) & 27
         & 64.6({\color{higher}+0.7}) & 28
         & 72.1({\color{higher}+0.1}) & 23 \\
         
         & \shine 
         & \bf60.3({\color{higher}+1.4}) & 142
         & \bf65.5({\color{higher}+1.6}) & 150
         & \bf73.1({\color{higher}+1.1}) & 81\\
		
	\hline
 
         \multirow{3}{*}{\rotatebox[origin=c]{90}{LLM}}
         
         & H-CLIP
         & 55.8({\color{lower}-3.1}) & 2
         & 60.1({\color{lower}-3.8}) & 2
         & 66.9({\color{lower}-5.1}) & 1 \\
         
         & CHiLS 
         & 61.1({\color{higher}+2.2}) & 26
         & 66.1({\color{higher}+2.2}) & 27
         & 73.4({\color{higher}+1.4}) & 23\\
         
         & \shine 
         & \bf61.6({\color{higher}+2.7}) & 141
         & \bf66.7({\color{higher}+2.8}) & 149
         & \bf73.6({\color{higher}+1.6}) & 81\\
         \bottomrule
    \end{tabular}
    \lesspace
    \lbltab{cls_comp}
\end{table}

\myparagraph{ImageNet-1k Benchmark.}
In \reftab{cls_comp}, we compare methods on
ImageNet in terms of accuracy and frames-per-second (FPS). We observe that our approach consistently outperforms 
related
methods. Comparing to the 
baseline that only uses class names, \shine improves its performance by an average of \textbf{\color{higher}+1.2\%} and \textbf{\color{higher}+2.4\%} across different model sizes using WordNet and LLM-generated hierarchies, {respectively}. 
Note that both \chils and \hclip introduce significant computational overheads due to their \textit{search-on-the-fly} mechanism, resulting in a considerable decrease in inference speed. Consequently, this limits their scalability to detection tasks that necessitate per-region proposal inference for each image. For example, when processing detection results for \textit{one} image with 300 region proposals, the overhead caused by \chils and \hclip would increase by $\approx$300$\times$. In contrast, \shine maintains the same inference speed as the baseline, 
preserving its scalability.

\begin{table}
    \centering
    \small
    \tablestyle{4.0pt}{1}

    \caption{
    BREEDS-structured~\cite{santurkar2020breeds} ImageNet-1k zero-shot classification (with varying granularity). All methods use the BREED hierarchy and use CLIP ViT-B/16. Top-1 accuracy (\%) reported.
    }
    \begin{tabular}{clllll}
        \toprule
        
        \multicolumn{1}{c}{Level} & \multicolumn{1}{c}{\# Classes} & \multicolumn{1}{c}{CLIP} & \multicolumn{1}{c}{H-CLIP~\cite{ge2023improving}} & \multicolumn{1}{c}{CHiLS~\cite{novack2023chils}} & \multicolumn{1}{c}{\shine} \\
        
        \midrule
        
        L1 & 10  & 56.2 & 67.9 ({\color{higher}+11.7})  & \bf73.8 ({\color{higher}+17.6})& 50.4({\color{lower}-5.8})\\
        L2 & 29  & 56.8 & \bf69.3 ({\color{higher}+12.5})  & 67.2 ({\color{higher}+10.4})& 60.9({\color{higher}+4.1})\\
        L3 & 128 & 43.3 & \bf62.4 ({\color{higher}+19.1}) & 62.2 ({\color{higher}+18.9})& 54.7({\color{higher}+11.4})\\
        L4 & 466 & 55.2 & 69.6 ({\color{higher}+14.4}) & 70.1 ({\color{higher}+14.9})& \bf70.3({\color{higher}+15.1})\\
        L5 & 591 & 62.4 & 65.9 ({\color{higher}+3.5})  & 64.5 ({\color{higher}+2.1})& \bf69.1({\color{higher}+6.7})\\
        L6 & 98  & 73.1 & 75.4 ({\color{higher}+2.3})  & 73.5 ({\color{higher}+0.4})& \bf78.9({\color{higher}+5.8})\\
        
        \bottomrule
    \end{tabular}
    \lesspace
    \lbltab{cls_breed_comp}
\end{table}

\myparagraph{BREEDS ImageNet Benchmark.} 
Next, we analyze different granularity levels within ImageNet as organized by BREEDS~\cite{santurkar2020breeds}.
In \reftab{cls_breed_comp}, we observe that \chils and \hclip surpass \shine at coarser granularity levels (L1 to L3). This 
is largely attributed to the BREEDS-modified hierarchy, where specific sub-classes 
in the hierarchy precisely correspond to the objects present in the test data.
Yet, our method yields more substantial performance improvements at finer granularity levels (L4 to L6). Overall, the 
performance gains exhibited
by all three 
methods underscore the benefits of using hierarchy information for improving 
open-vocabulary performance across granularities.

\section{Conclusion}
\lblsec{conclusion}
Given the importance of the vocabulary in open-vocabulary object detection, the robustness to varying granularities becomes critical for off-the-shelf deployment of \ovod models. Our preliminary investigations uncovered notable performance variability in existing \ovod detectors across different vocabulary granularities. To address this, we 
introduced \shine, a novel method that utilizes semantic knowledge from hierarchies to build \textit{nexus}-based classifiers. \shine is training-free and can be seamlessly integrated with any \ovod detector, maintaining linear complexity relative to 
the number of classes.
We show that \shine yields consistent improvements over baseline detectors across granularities with ground truth and LLM-generated hierarchies. We also extend \shine to open-vocabulary classification and achieve notable gains on ImageNet-1k~\cite{deng2009imagenet}.


\vspace{1.5 mm}
\noindent\small{\textbf{Acknowledgements.}~E.R. is supported by MUR PNRR project FAIR - Future AI Research (PE00000013), funded by NextGenerationEU and EU projects SPRING (No. 871245) and ELIAS (No. 01120237). M.L. is supported by the PRIN project LEGO-AI (Prot. 2020TA3K9N).
We thank Diane Larlus and Yannis Kalantidis for their helpful suggestions.
M.L. thanks Zhun Zhong and Margherita Potrich for their constant
support.
}

{
    \small
    \bibliographystyle{ieeenat_fullname}
    \bibliography{main}

\begin{thebibliography}{73}
\providecommand{\natexlab}[1]{#1}
\providecommand{\url}[1]{\texttt{#1}}
\expandafter\ifx\csname urlstyle\endcsname\relax
  \providecommand{\doi}[1]{doi: #1}\else
  \providecommand{\doi}{doi: \begingroup \urlstyle{rm}\Url}\fi

\bibitem[Arandjelovi{\'c} et~al.(2023)Arandjelovi{\'c}, Andonian, Mensch, H{\'e}naff, Alayrac, and Zisserman]{arandjelovic2023three}
Relja Arandjelovi{\'c}, Alex Andonian, Arthur Mensch, Olivier~J. H{\'e}naff, Jean-Baptiste Alayrac, and Andrew Zisserman.
\newblock {Three ways to Improve Feature Alignment for Open Vocabulary Detection}.
\newblock arXiv:2303.13518, 2023.

\bibitem[Barz and Denzler(2019)]{barz2019hierarchy}
Bj{\"o}rn Barz and Joachim Denzler.
\newblock {Hierarchy-based Image Embeddings for Semantic Image Retrieval}.
\newblock In \emph{WACV}, 2019.

\bibitem[Bertinetto et~al.(2020)Bertinetto, Mueller, Tertikas, Samangooei, and Lord]{bertinetto2020making}
Luca Bertinetto, Romain Mueller, Konstantinos Tertikas, Sina Samangooei, and Nicholas~A. Lord.
\newblock {Making Better Mistakes: Leveraging Class Hierarchies with Deep Networks}.
\newblock In \emph{CVPR}, 2020.

\bibitem[Brown et~al.(2020)Brown, Mann, Ryder, Subbiah, Kaplan, Dhariwal, Neelakantan, Shyam, Sastry, Askell, et~al.]{brown2020language}
Tom Brown, Benjamin Mann, Nick Ryder, Melanie Subbiah, Jared~D. Kaplan, Prafulla Dhariwal, Arvind Neelakantan, Pranav Shyam, Girish Sastry, Amanda Askell, et~al.
\newblock {Language Models are Few-Shot Learners}.
\newblock In \emph{NeurIPS}, 2020.

\bibitem[Carion et~al.(2020)Carion, Massa, Synnaeve, Usunier, Kirillov, and Zagoruyko]{carion2020end}
Nicolas Carion, Francisco Massa, Gabriel Synnaeve, Nicolas Usunier, Alexander Kirillov, and Sergey Zagoruyko.
\newblock {End-to-end object detection with transformers}.
\newblock In \emph{ECCV}, 2020.

\bibitem[Cole et~al.(2022)Cole, Wilber, Van~Horn, Yang, Fornoni, Perona, Belongie, Howard, and Aodha]{cole2022label}
Elijah Cole, Kimberly Wilber, Grant Van~Horn, Xuan Yang, Marco Fornoni, Pietro Perona, Serge Belongie, Andrew Howard, and Oisin~Mac Aodha.
\newblock {On Label Granularity and Object Localization}.
\newblock In \emph{ECCV}, 2022.

\bibitem[Deng et~al.(2009)Deng, Dong, Socher, Li, Li, and Fei-Fei]{deng2009imagenet}
Jia Deng, Wei Dong, Richard Socher, Li-Jia Li, Kai Li, and Li Fei-Fei.
\newblock {ImageNet: a Large-Scale Hierarchical Image Database}.
\newblock In \emph{CVPR}, 2009.

\bibitem[Deng et~al.(2010)Deng, Berg, Li, and Fei-Fei]{deng2010does}
Jia Deng, Alexander~C. Berg, Kai Li, and Li Fei-Fei.
\newblock {What Does Classifying more than 10,000 Image Categories Tell Us?}
\newblock In \emph{ECCV}, 2010.

\bibitem[Du et~al.(2022)Du, Wei, Zhang, Shi, Gao, and Li]{du2022learning}
Yu Du, Fangyun Wei, Zihe Zhang, Miaojing Shi, Yue Gao, and Guoqi Li.
\newblock {Learning to Prompt for Open-Vocabulary Object Detection with Vision-Language Model}.
\newblock In \emph{CVPR}, 2022.

\bibitem[Fan et~al.(2020)Fan, Zhuo, Tang, and Tai]{fan2020few}
Qi Fan, Wei Zhuo, Chi-Keung Tang, and Yu-Wing Tai.
\newblock {Few-Shot Object Detection with Attention-RPN and Multi-Relation Detector}.
\newblock In \emph{CVPR}, 2020.

\bibitem[Fellbaum(1998)]{fellbaum1998wordnet}
Christiane Fellbaum.
\newblock \emph{{WordNet: an Electronic Lexical Database}}.
\newblock MIT Press, 1998.

\bibitem[Feng et~al.(2022)Feng, Zhong, Jie, Chu, Ren, Wei, Xie, and Ma]{feng2022promptdet}
Chengjian Feng, Yujie Zhong, Zequn Jie, Xiangxiang Chu, Haibing Ren, Xiaolin Wei, Weidi Xie, and Lin Ma.
\newblock {PromptDet: Towards Open-vocabulary Detection using Uncurated Images}.
\newblock In \emph{ECCV}, 2022.

\bibitem[Frome et~al.(2013)Frome, Corrado, Shlens, Bengio, Dean, Ranzato, and Mikolov]{frome2013devise}
Andrea Frome, Greg~S. Corrado, Jon Shlens, Samy Bengio, Jeff Dean, Marc'Aurelio Ranzato, and Tomas Mikolov.
\newblock {DeViSE: A Deep Visual-Semantic Embedding Model}.
\newblock In \emph{NeurIPS}, 2013.

\bibitem[Ge et~al.(2023)Ge, Ren, Gallagher, Wang, Yang, Adam, Itti, Lakshminarayanan, and Zhao]{ge2023improving}
Yunhao Ge, Jie Ren, Andrew Gallagher, Yuxiao Wang, Ming-Hsuan Yang, Hartwig Adam, Laurent Itti, Balaji Lakshminarayanan, and Jiaping Zhao.
\newblock {Improving Zero-shot Generalization and Robustness of Multi-modal Models}.
\newblock In \emph{CVPR}, 2023.

\bibitem[Gewers et~al.(2021)Gewers, Ferreira, {de Arruda}, Silva, Comin, Amancio, and Costa]{gewers2021principal}
Felipe~L. Gewers, Gustavo~R. Ferreira, Henrique~F. {de Arruda}, Filipi~N. Silva, Cesar~H. Comin, Diego~R. Amancio, and Luciano da~F. Costa.
\newblock {Principal Component Analysis: A Natural Approach to Data Exploration}.
\newblock \emph{ACM Computing Surveys (CSUR)}, 54\penalty0 (4):\penalty0 1--34, 2021.

\bibitem[Goodman(2001)]{goodman2001classes}
Joshua Goodman.
\newblock {Classes for Fast Maximum Entropy Training}.
\newblock In \emph{ICASSP}, 2001.

\bibitem[Gu et~al.(2023)Gu, Han, Chen, Beirami, He, Zhang, Liao, Qin, Tresp, and Torr]{gu2023systematic}
Jindong Gu, Zhen Han, Shuo Chen, Ahmad Beirami, Bailan He, Gengyuan Zhang, Ruotong Liao, Yao Qin, Volker Tresp, and Philip Torr.
\newblock {A Systematic Survey of Prompt Engineering on Vision-Language Foundation Models}.
\newblock arXiv:2307.12980, 2023.

\bibitem[Gu et~al.(2022)Gu, Lin, Kuo, and Cui]{gu2021open}
Xiuye Gu, Tsung-Yi Lin, Weicheng Kuo, and Yin Cui.
\newblock {Open-vocabulary Object Detection via Vision and Language Knowledge Distillation}.
\newblock In \emph{ICLR}, 2022.

\bibitem[Gupta et~al.(2019)Gupta, Dollar, and Girshick]{gupta2019lvis}
Agrim Gupta, Piotr Dollar, and Ross Girshick.
\newblock {LVIS: A Dataset for Large Vocabulary Instance Segmentation}.
\newblock In \emph{CVPR}, 2019.

\bibitem[Hamamci et~al.(2023)Hamamci, Er, Simsar, Sekuboyina, Gundogar, Stadlinger, Mehl, and Menze]{hamamci2023diffusion}
Ibrahim~Ethem Hamamci, Sezgin Er, Enis Simsar, Anjany Sekuboyina, Mustafa Gundogar, Bernd Stadlinger, Albert Mehl, and Bjoern Menze.
\newblock {Diffusion-Based Hierarchical Multi-Label Object Detection to Analyze Panoramic Dental X-rays}.
\newblock In \emph{MICCAI}, 2023.

\bibitem[He et~al.(2016)He, Zhang, Ren, and Sun]{he2016deep}
Kaiming He, Xiangyu Zhang, Shaoqing Ren, and Jian Sun.
\newblock {Deep Residual Learning for Image Recognition}.
\newblock In \emph{CVPR}, 2016.

\bibitem[Jia et~al.(2021)Jia, Yang, Xia, Chen, Parekh, Pham, Le, Sung, Li, and Duerig]{jia2021scaling}
Chao Jia, Yinfei Yang, Ye Xia, Yi-Ting Chen, Zarana Parekh, Hieu Pham, Quoc Le, Yun-Hsuan Sung, Zhen Li, and Tom Duerig.
\newblock {Scaling up visual and vision-language representation learning with noisy text supervision}.
\newblock In \emph{ICML}, 2021.

\bibitem[Kaul et~al.(2023)Kaul, Xie, and Zisserman]{kaul2023multi}
Prannay Kaul, Weidi Xie, and Andrew Zisserman.
\newblock {Multi-Modal Classifiers for Open-Vocabulary Object Detection}.
\newblock In \emph{ICML}, 2023.

\bibitem[Khattak et~al.(2023)Khattak, Rasheed, Maaz, Khan, and Khan]{khattak2023maple}
Muhammad~Uzair Khattak, Hanoona Rasheed, Muhammad Maaz, Salman Khan, and Fahad~Shahbaz Khan.
\newblock {MaPLe: Multi-modal Prompt Learning}.
\newblock In \emph{CVPR}, 2023.

\bibitem[Knight(2002)]{knight2002safety}
John~C. Knight.
\newblock {Safety Critical Systems: Challenges and Directions}.
\newblock In \emph{ICSE}, 2002.

\bibitem[Kuo et~al.(2023)Kuo, Cui, Gu, Piergiovanni, and Angelova]{kuo2022f}
Weicheng Kuo, Yin Cui, Xiuye Gu, AJ Piergiovanni, and Anelia Angelova.
\newblock {F-VLM: Open-Vocabulary Object Detection upon Frozen Vision and Language Models}.
\newblock In \emph{ICLR}, 2023.

\bibitem[Kuznetsova et~al.(2020)Kuznetsova, Rom, Alldrin, Uijlings, Krasin, Pont-Tuset, Kamali, Popov, Malloci, Kolesnikov, Duerig, and Ferrari]{kuznetsova2020open}
Alina Kuznetsova, Hassan Rom, Neil Alldrin, Jasper Uijlings, Ivan Krasin, Jordi Pont-Tuset, Shahab Kamali, Stefan Popov, Matteo Malloci, Alexander Kolesnikov, Tom Duerig, and Vittorio Ferrari.
\newblock {The Open Images Dataset V4}.
\newblock \emph{IJCV}, 128:\penalty0 1956--1981, 2020.

\bibitem[Li and Jain(1998)]{li1998classification}
Yong~H Li and Anil~K. Jain.
\newblock {Classification of Text Documents}.
\newblock \emph{The Computer Journal}, 41\penalty0 (8):\penalty0 537--546, 1998.

\bibitem[Lin et~al.(2023{\natexlab{a}})Lin, Sun, Jiang, Luo, Qu, Haffari, Yuan, and Cai]{lin2022learning}
Chuang Lin, Peize Sun, Yi Jiang, Ping Luo, Lizhen Qu, Gholamreza Haffari, Zehuan Yuan, and Jianfei Cai.
\newblock {Interpreting Word Embeddings with Eigenvector Analysis}.
\newblock In \emph{ICLR}, 2023{\natexlab{a}}.

\bibitem[Lin et~al.(2014)Lin, Maire, Belongie, Hays, Perona, Ramanan, Doll{\'a}r, and Zitnick]{lin2014microsoft}
Tsung-Yi Lin, Michael Maire, Serge Belongie, James Hays, Pietro Perona, Deva Ramanan, Piotr Doll{\'a}r, and C.~Lawrence Zitnick.
\newblock {Microsoft COCO: Common Objects in Context}.
\newblock In \emph{ECCV}, 2014.

\bibitem[Lin et~al.(2023{\natexlab{b}})Lin, Karlinsky, Shvetsova, Possegger, Kozinski, Panda, Feris, Kuehne, and Bischof]{lin2023match}
Wei Lin, Leonid Karlinsky, Nina Shvetsova, Horst Possegger, Mateusz Kozinski, Rameswar Panda, Rogerio Feris, Hilde Kuehne, and Horst Bischof.
\newblock {MAtch, eXpand and Improve: Unsupervised Finetuning for Zero-Shot Action Recognition with Language Knowledge}.
\newblock In \emph{ICCV}, 2023{\natexlab{b}}.

\bibitem[Liu et~al.(2021)Liu, Lin, Cao, Hu, Wei, Zhang, Lin, and Guo]{liu2021swin}
Ze Liu, Yutong Lin, Yue Cao, Han Hu, Yixuan Wei, Zheng Zhang, Stephen Lin, and Baining Guo.
\newblock {Swin Transformer: Hierarchical Vision Transformer using Shifted Windows}.
\newblock In \emph{ICCV}, 2021.

\bibitem[Ma et~al.(2023)Ma, Jiang, Wen, Yuan, and Qi]{ma2024codet}
Chuofan Ma, Yi Jiang, Xin Wen, Zehuan Yuan, and Xiaojuan Qi.
\newblock {CoDet: Co-Occurrence Guided Region-Word Alignment for Open-Vocabulary Object Detection}.
\newblock In \emph{NeurIPS}, 2023.

\bibitem[Mart{\'\i}nez-D{\'\i}az and Soriguera(2018)]{martinez2018autonomous}
Margarita Mart{\'\i}nez-D{\'\i}az and Francesc Soriguera.
\newblock {Autonomous Vehicles: Theoretical and Practical Challenges}.
\newblock \emph{Transportation Research Procedia}, 33:\penalty0 275--282, 2018.

\bibitem[Menon and Vondrick(2023)]{menon2022visual}
Sachit Menon and Carl Vondrick.
\newblock {Visual Classification via Description from Large Language Models}.
\newblock In \emph{ICLR}, 2023.

\bibitem[Minderer et~al.(2023)Minderer, Gritsenko, and Houlsby]{minderer2023scaling}
Matthias Minderer, Alexey Gritsenko, and Neil Houlsby.
\newblock {Scaling Open-Vocabulary Object Detection}.
\newblock In \emph{NeurIPS}, 2023.

\bibitem[Morin and Bengio(2005)]{morin2005hierarchical}
Frederic Morin and Yoshua Bengio.
\newblock {Hierarchical Probabilistic Neural Network Language Model}.
\newblock In \emph{{International Workshop on Artificial Intelligence and Statistics}}, 2005.

\bibitem[Novack et~al.(2023)Novack, McAuley, Lipton, and Garg]{novack2023chils}
Zachary Novack, Julian McAuley, Zachary~Chase Lipton, and Saurabh Garg.
\newblock {CHiLS: Zero-Shot Image Classification with Hierarchical Label Sets}.
\newblock In \emph{ICML}, 2023.

\bibitem[{OpenAI}(2022)]{chatgpt}
{OpenAI}.
\newblock {ChatGPT: A Large-Scale GPT-3.5-Based Model}.
\newblock \url{https://openai.com/blog/chatgpt}, 2022.

\bibitem[Parashar et~al.(2023)Parashar, Lin, Li, and Kong]{parashar2023prompting}
Shubham Parashar, Zhiqiu Lin, Yanan Li, and Shu Kong.
\newblock {Prompting Scientific Names for Zero-Shot Species Recognition}.
\newblock In \emph{EMNLP}, 2023.

\bibitem[Pratt et~al.(2023)Pratt, Covert, Liu, and Farhadi]{pratt2023does}
Sarah Pratt, Ian Covert, Rosanne Liu, and Ali Farhadi.
\newblock {What Does a Platypus Look Like? Generating Customized Prompts for Zero-shot Image Classification}.
\newblock In \emph{ICCV}, 2023.

\bibitem[Radford et~al.(2021)Radford, Kim, Hallacy, Ramesh, Goh, Agarwal, Sastry, Askell, Mishkin, Clark, Krueger, and Sutskever]{radford2021learning}
Alec Radford, Jong~Wook Kim, Chris Hallacy, Aditya Ramesh, Gabriel Goh, Sandhini Agarwal, Girish Sastry, Amanda Askell, Pamela Mishkin, Jack Clark, Gretchen Krueger, and Ilya Sutskever.
\newblock {Learning Transferable Visual Models From Natural Language Supervision}.
\newblock In \emph{ICML}, 2021.

\bibitem[Ren et~al.(2023)Ren, Zhang, Zhu, Zhang, Zheng, Li, Smola, and Sun]{ren2023prompt}
Shuhuai Ren, Aston Zhang, Yi Zhu, Shuai Zhang, Shuai Zheng, Mu Li, Alex Smola, and Xu Sun.
\newblock {Prompt Pre-Training with Twenty-Thousand Classes for Open-Vocabulary Visual Recognition}.
\newblock In \emph{NeurIPS}, 2023.

\bibitem[Ridnik et~al.(2021)Ridnik, Ben-Baruch, Noy, and Zelnik-Manor]{ridnik2021imagenet}
Tal Ridnik, Emanuel Ben-Baruch, Asaf Noy, and Lihi Zelnik-Manor.
\newblock {ImageNet-21K Pretraining for the Masses}.
\newblock In \emph{NeurIPS}, 2021.

\bibitem[Roth et~al.(2023)Roth, Kim, Koepke, Vinyals, Schmid, and Akata]{roth2023waffling}
Karsten Roth, Jae~Myung Kim, A.~Sophia Koepke, Oriol Vinyals, Cordelia Schmid, and Zeynep Akata.
\newblock {Waffling around for Performance: Visual Classification with Random Words and Broad Concepts}.
\newblock In \emph{ICCV}, 2023.

\bibitem[Ruggiero et~al.(2015)Ruggiero, Gordon, Orrell, Bailly, Bourgoin, Brusca, Cavalier-Smith, Guiry, and Kirk]{ruggiero2015higher}
Michael~A. Ruggiero, Dennis~P. Gordon, Thomas~M. Orrell, Nicolas Bailly, Thierry Bourgoin, Richard~C. Brusca, Thomas Cavalier-Smith, Michael~D. Guiry, and Paul~M. Kirk.
\newblock {A Higher Level Classification of All Living Organisms}.
\newblock \emph{{PLOS ONE}}, 10\penalty0 (4):\penalty0 e0119248, 2015.

\bibitem[Ruiz and Srinivasan(2002)]{ruiz2002hierarchical}
Miguel~E. Ruiz and Padmini Srinivasan.
\newblock {Hierarchical Text Categorization Using Neural Networks}.
\newblock \emph{Information retrieval}, 5\penalty0 (1):\penalty0 87--118, 2002.

\bibitem[Santurkar et~al.(2021)Santurkar, Tsipras, and Madry]{santurkar2020breeds}
Shibani Santurkar, Dimitris Tsipras, and Aleksander Madry.
\newblock {BREEDS: Benchmarks for Subpopulation Shift}.
\newblock In \emph{ICLR}, 2021.

\bibitem[Sharma et~al.(2018)Sharma, Ding, Goodman, and Soricut]{sharma2018conceptual}
Piyush Sharma, Nan Ding, Sebastian Goodman, and Radu Soricut.
\newblock {Conceptual Captions: A Cleaned, Hypernymed, Image Alt-text Dataset For Automatic Image Captioning}.
\newblock In \emph{ACL}, 2018.

\bibitem[Shin et~al.(2018)Shin, Madotto, and Fung]{shin2018interpreting}
Jamin Shin, Andrea Madotto, and Pascale Fung.
\newblock {Interpreting Word Embeddings with Eigenvector Analysis}.
\newblock In \emph{NeurIPS, IRASL workshop}, 2018.

\bibitem[Shin et~al.(2020)Shin, Kim, Kim, and Kim]{shin2020hierarchical}
Su-Jin Shin, Seyeob Kim, Youngjung Kim, and Sungho Kim.
\newblock {Hierarchical Multi-Label Object Detection Framework for Remote Sensing Images}.
\newblock \emph{Remote Sensing}, 12\penalty0 (17):\penalty0 2734, 2020.

\bibitem[Shu et~al.(2022)Shu, Nie, Huang, Yu, Goldstein, Anandkumar, and Xiao]{shu2022test}
Manli Shu, Weili Nie, De-An Huang, Zhiding Yu, Tom Goldstein, Anima Anandkumar, and Chaowei Xiao.
\newblock {Test-Time Prompt Tuning for Zero-Shot Generalization in Vision-Language Models}.
\newblock In \emph{NeurIPS}, 2022.

\bibitem[Silla and Freitas(2011)]{silla2011survey}
Carlos~N. Silla and Alex~A. Freitas.
\newblock {A Survey of Hierarchical Classification across Different Application Domains}.
\newblock \emph{{Data Mining and Knowledge Discovery}}, 22:\penalty0 31--72, 2011.

\bibitem[Tan et~al.(2021)Tan, Xu, and Shen]{tan2021survey}
Chufeng Tan, Xing Xu, and Fumin Shen.
\newblock {A Survey of Zero Shot Detection: Methods and Applications}.
\newblock \emph{Cognitive Robotics}, 1:\penalty0 159--167, 2021.

\bibitem[Van~Horn et~al.(2018)Van~Horn, Mac~Aodha, Song, Cui, Sun, Shepard, Adam, Perona, and Belongie]{van2018inaturalist}
Grant Van~Horn, Oisin Mac~Aodha, Yang Song, Yin Cui, Chen Sun, Alex Shepard, Hartwig Adam, Pietro Perona, and Serge Belongie.
\newblock {The iNaturalist Species Classification and Detection Dataset}.
\newblock In \emph{CVPR}, 2018.

\bibitem[Wah et~al.(2011)Wah, Branson, Welinder, Perona, and Belongie]{wah2011caltech}
Catherine Wah, Steve Branson, Peter Welinder, Pietro Perona, and Serge Belongie.
\newblock {The {Caltech-UCSD} Birds-200-2011 Dataset}.
\newblock Technical Report CNS-TR-2011-001, {California Institute of Technology}, 2011.

\bibitem[Wang et~al.(2023)Wang, Sun, Li, and Yang]{wang2023transhp}
Wenhao Wang, Yifan Sun, Wei Li, and Yi Yang.
\newblock {TransHP: Image Classification with Hierarchical Prompting}.
\newblock In \emph{NeurIPS}, 2023.

\bibitem[Wu et~al.(2005)Wu, Zhang, and Honavar]{wu2005learning}
Feihong Wu, Jun Zhang, and Vasant Honavar.
\newblock {Learning Classifiers Using Hierarchically Structured Class Taxonomies}.
\newblock In \emph{SARA}, 2005.

\bibitem[Wu et~al.(2024)Wu, Li, Yuan, Ding, Yang, Li, Zhang, Tong, Jiang, Ghanem, and Tao]{WuPAMI23TowardsOpenVocabularyLearning}
Jianzong Wu, Xiangtai Li, Shilin Xu~Haobo Yuan, Henghui Ding, Yibo Yang, Xia Li, Jiangning Zhang, Yunhai Tong, Xudong Jiang, Bernard Ghanem, and Dacheng Tao.
\newblock {Towards Open Vocabulary Learning: A Survey}.
\newblock \emph{IEEE TPAMI}, 2024.

\bibitem[Wu et~al.(2023{\natexlab{a}})Wu, Zhang, Jin, Liu, and Loy]{wu2023aligning}
Size Wu, Wenwei Zhang, Sheng Jin, Wentao Liu, and Chen~Change Loy.
\newblock {Aligning Bag of Regions for Open-Vocabulary Object Detection}.
\newblock In \emph{CVPR}, 2023{\natexlab{a}}.

\bibitem[Wu et~al.(2023{\natexlab{b}})Wu, Zhu, Zhao, and Li]{wu2023cora}
Xiaoshi Wu, Feng Zhu, Rui Zhao, and Hongsheng Li.
\newblock {CORA: Adapting CLIP for Open-Vocabulary Detection with Region Prompting and Anchor Pre-Matching}.
\newblock In \emph{CVPR}, 2023{\natexlab{b}}.

\bibitem[Yan et~al.(2023)Yan, Wang, Zhong, Dong, He, Lu, Wang, Shang, and McAuley]{yan2023learning}
An Yan, Yu Wang, Yiwu Zhong, Chengyu Dong, Zexue He, Yujie Lu, William~Yang Wang, Jingbo Shang, and Julian McAuley.
\newblock {Learning Concise and Descriptive Attributes for Visual Recognition}.
\newblock In \emph{ICCV}, 2023.

\bibitem[Yao et~al.(2022)Yao, Han, Wen, Liang, Xu, Zhang, Li, Xu, and Xu]{yao2022detclip}
Lewei Yao, Jianhua Han, Youpeng Wen, Xiaodan Liang, Dan Xu, Wei Zhang, Zhenguo Li, Chunjing Xu, and Hang Xu.
\newblock {DetCLIP: Dictionary-Enriched Visual-Concept Paralleled Pre-training for Open-world Detection}.
\newblock In \emph{NeurIPS}, 2022.

\bibitem[Zang et~al.(2022)Zang, Li, Zhou, Huang, and Loy]{zang2022open}
Yuhang Zang, Wei Li, Kaiyang Zhou, Chen Huang, and Chen~Change Loy.
\newblock {Open-Vocabulary DETR with Conditional Matching}.
\newblock In \emph{ECCV}, 2022.

\bibitem[Zareian et~al.(2021)Zareian, Rosa, Hu, and Chang]{zareian2021open}
Alireza Zareian, Kevin~Dela Rosa, Derek~Hao Hu, and Shih-Fu Chang.
\newblock {Open-Vocabulary Object Detection Using Captions}.
\newblock In \emph{CVPR}, 2021.

\bibitem[Zhang et~al.(2023{\natexlab{a}})Zhang, Ren, Gu, and Tresp]{zhang2023multi}
Gengyuan Zhang, Jisen Ren, Jindong Gu, and Volker Tresp.
\newblock {Multi-event Video-Text Retrieval}.
\newblock In \emph{ICCV}, 2023{\natexlab{a}}.

\bibitem[Zhang et~al.(2023{\natexlab{b}})Zhang, Li, Zou, Liu, Li, Yang, and Zhang]{zhang2023simple}
Hao Zhang, Feng Li, Xueyan Zou, Shilong Liu, Chunyuan Li, Jianwei Yang, and Lei Zhang.
\newblock {A Simple Framework for Open-Vocabulary Segmentation and Detection}.
\newblock In \emph{ICCV}, 2023{\natexlab{b}}.

\bibitem[Zhong et~al.(2022)Zhong, Yang, Zhang, Li, Codella, Li, Zhou, Dai, Yuan, Li, and Gao]{zhong2022regionclip}
Yiwu Zhong, Jianwei Yang, Pengchuan Zhang, Chunyuan Li, Noel Codella, Liunian~Harold Li, Luowei Zhou, Xiyang Dai, Lu Yuan, Yin Li, and Jianfeng Gao.
\newblock {RegionCLIP: Region-based Language-Image Pretraining}.
\newblock In \emph{CVPR}, 2022.

\bibitem[Zhou et~al.(2022{\natexlab{a}})Zhou, Yang, Loy, and Liu]{zhou2022conditional}
Kaiyang Zhou, Jingkang Yang, Chen~Change Loy, and Ziwei Liu.
\newblock {Conditional Prompt Learning for Vision-Language Models}.
\newblock In \emph{CVPR}, 2022{\natexlab{a}}.

\bibitem[Zhou et~al.(2022{\natexlab{b}})Zhou, Yang, Loy, and Liu]{zhou2022learning}
Kaiyang Zhou, Jingkang Yang, Chen~Change Loy, and Ziwei Liu.
\newblock {Learning to Prompt for Vision-Language Models}.
\newblock \emph{IJCV}, 130\penalty0 (7):\penalty0 2337–2348, 2022{\natexlab{b}}.

\bibitem[Zhou et~al.(2021)Zhou, Koltun, and Kr{\"a}henb{\"u}hl]{zhou2021probabilistic}
Xingyi Zhou, Vladlen Koltun, and Philipp Kr{\"a}henb{\"u}hl.
\newblock {Probabilistic two-stage Detection}.
\newblock arXiv:2103.07461, 2021.

\bibitem[Zhou et~al.(2022{\natexlab{c}})Zhou, Girdhar, Joulin, Kr{\"a}henb{\"u}hl, and Misra]{zhou2022detecting}
Xingyi Zhou, Rohit Girdhar, Armand Joulin, Philipp Kr{\"a}henb{\"u}hl, and Ishan Misra.
\newblock {Detecting Twenty-thousand Classes using Image-level Supervision}.
\newblock In \emph{ECCV}, 2022{\natexlab{c}}.

\bibitem[Zhu and Chen(2023)]{zhu2023survey}
Chaoyang Zhu and Long Chen.
\newblock {A Survey on Open-Vocabulary Detection and Segmentation: Past, Present, and Future}.
\newblock arXiv:2307.09220, 2023.

\end{thebibliography}
}

\clearpage
\maketitlesupplementary
\section*{Appendix}
\appendix

In this appendix, we begin by detailing the detection datasets in \refsupp{supp_data}. 
Then,
\refsupp{supp_hier} delves into the process of synthetic semantic hierarchy generation using \llm{s}, providing the \llm{} prompts and a thorough summary of the generated hierarchies' statistics. 
We provide in \refsupp{supp_impl} additional implementation specifics of \shine. 
We present in \refsupp{supp_expt_ablation_fsod} an extended analysis of \shine's components on the \fsod dataset. \refsupp{supp_expt_more_swinb} and \refsupp{supp_expt_stat} extend the main detection experiments, offering comprehensive summary statistics. \refsupp{supp_expt_cocolvis} includes additional investigation of \shine{'s performance} on the COCO and LVIS datasets. Finally, \refsupp{supp_qua} showcases some qualitative detection results. 
{Our code is publicly available at \href{https://github.com/naver/shine}{https://github.com/naver/shine}}.

\section{Dataset Details}
\lblsec{supp_data}
\begin{table}[!h]
    \centering
    \tablestyle{4.5pt}{1.2}
    \caption{
    Summary of datasets. 
    \fsod~\cite{fan2020few} and \inat~\cite{cole2022label} offer three and six levels of label hierarchies, respectively, with varying semantic granularity. This feature allows evaluating models on these datasets at different granularity level in our experiments. To evaluate, for \fsod, we use its official test split, and for \inat, we use the combined test and validation splits.
    }
    \lesspace
    \begin{tabular}{l|ccc|cccccc}
        \toprule
         \multicolumn{1}{c}{}& \multicolumn{3}{c}{\fsod (test split)} 
         & \multicolumn{6}{c}{\inat (test + val splits)} \\

        \hline
        
         \# of Levels 
         & \multicolumn{3}{c|}{3} 
         & \multicolumn{6}{c}{6} \\
         
         \# of Classes
         & L3 & L2 & L1
         & L6 & L5 & L4 & L3 & L2 & L1 \\
         
         per Level 
         & 200 & 46 & 15
         & 500 & 317 & 184 & 64 & 18 & 5 \\

         \# of Images
         & \multicolumn{3}{c|}{14152} 
         & \multicolumn{6}{c}{25000} \\

         \# of BBoxes
         & \multicolumn{3}{c|}{35102} 
         & \multicolumn{6}{c}{25000} \\

         \bottomrule
    \end{tabular}
    \lesspace
    \lbltab{full_dataset_statistics}
\end{table}

In \reftab{full_dataset_statistics}, we present a comprehensive summary of the two detection datasets used in our evaluation, \fsod~\cite{fan2020few} and \inat~\cite{cole2022label}, under the cross-dataset transfer open-vocabulary evaluation protocol. Given that the original \fsod dataset~\cite{fan2020few} provides only a two-level hierarchy, we manually constructed the L1 level of the label hierarchy (the most abstract one) for a more comprehensive evaluation, under which we grouped the categories of level L2 (which corresponds to the upper level category in the original label space). The L1 level consists of the following 15 label categories:\\

\mytexttt{\{"liquid", "instrument", "food", "art", "plant", "component",  "animal", "body",  "wearable item", "infrastructure",  "vehicle", "furnishing", "fungi",  "equipment", "beauty product"\}} \\

For \fsod and \inat, we use their official test splits in our experiments, respectively. Notably, \fsod and \inat offer three and six levels of label hierarchies, respectively, each characterized by distinct semantic granularities. Specifically, for the same set of evaluation images and their associated box-label annotations, the actual \textit{label} used for evaluation can be mapped to different linked labels at each granularity level of the hierarchy. For example, in \fsod, a box region labeled as \mytexttt{"watermelon"} at the L3-level could be mapped to label \mytexttt{"fruit"} at the L2-level or \mytexttt{"food"} at the L1-level. This hierarchical approach to labeling facilitates the evaluation of these datasets at various granularity levels. 
{See the annotation files in our \href{https://github.com/naver/shine}{codebase}}.

\myparagraph{Evaluation level.}
During the evaluation, we consider only one hierarchy level at a time, where the label class vocabulary corresponding to the evaluation level of granularity serves as the target test (user-defined) vocabulary for both the methods being compared and our proposed \shine. This means that model has to assign labels solely from the given hierarchy level.

\section{Semantic Hierarchy Generation}
\lblsec{supp_hier}

\begin{table*}[!t]
    \centering
    \tablestyle{8.5pt}{1.1}
    \caption{
    Summary statistics of synthetic hierarchies generated by the \llm{} for our experiments. We present the number of label classes in the target vocabulary, the total number of generated super-categories and sub-categories, and the average number of generated super-categories and sub-categories per target class (CoI) for each dataset at each label vocabulary level. Additionally, links are provided to the experiments using these \llm-generated hierarchies. Note: N/A indicates that only one level of label vocabulary exists in the dataset. \dag: At the most abstract (coarsest) level L1 of \inat, all target classes belong to the single super-category Kingdom \texttt{"Animalia"}.
    }
    \lesspace
    \begin{tabular}{lccccccc}
         \toprule
         \multirow{2}{*}{Dataset} & Corresponding & Label Vocabulary & Number of 
         & \multicolumn{2}{c}{Number of Super-categories} & \multicolumn{2}{c}{Number of Sub-categories} \\
         
         & Experiments & Level & Label Classes & Total & Avg. per class & Total & Avg. per class\\
         
         \midrule

         \multirow{6}{*}{\inat~\cite{cole2022label}}
         & \multirow{6}{*}{
         \reftab{imprv_detic_swin}, \ref{tbl:imprv_vldet_codet_swinb}, \ref{tbl:comp_additional_baselines},
         \ref{tbl:imprv_detic_swinB_12only}, \ref{tbl:imprv_vldet_cora_rn50}
         }
         & L6 & 500 
         & 317 & 0.6 
         & 10909 & 21.8 \\

         &
         & L5 & 317 
         & 184 & 0.6 
         & 6675 & 21.1 \\

         &
         & L4 & 184 
         & 64 & 0.3 
         & 3102 & 16.9 \\
         
         &
         & L3 & 64 
         & 18 & 0.3 
         & 1018 & 16.7 \\

         &
         & L2 & 18 
         & 5 & 0.3 
         & 273 & 15.2 \\

         &
         & L1 & 5 
         & 1\dag & 0.2 
         & 63 & 12.6 \\

         & \reffig{expanded_comp}(a)
         & Mis-spe. L6 & 1966 
         & 7585 & 3.9 
         & 35151 & 17.9 \\
         \midrule

         \multirow{3}{*}{\fsod~\cite{fan2020few}}
         & \multirow{3}{*}{
         \reftab{imprv_detic_swin}, \ref{tbl:imprv_vldet_codet_swinb}, \ref{tbl:comp_additional_baselines},
         \ref{tbl:imprv_detic_swinB_12only}, \ref{tbl:imprv_vldet_cora_rn50}
         }
         & L3 & 200 
         & 1298 & 6.5 
         & 4140 & 20.7 \\

         &
         & L2 & 46 
         & 295 & 6.4 
         & 702 & 15.2 \\

         &
         & L1 & 15 
         & 92 & 6.1 
         & 239 & 15.9 \\

         & \reffig{expanded_comp}(b)
         & Mis-spe. L3 & 1570 
         & 8069 & 5.1 
         & 26684 & 17.0 \\

        \midrule
         COCO~\cite{lin2014microsoft}
         & \reftab{base_novel_comp}
         & N/A & 65
         & 395 & 6.1 
         & 1303 & 20.1 \\      

         \midrule
         LVIS~\cite{gupta2019lvis}
         & \reftab{base_novel_comp}
         & N/A & 1203
         & 6016 & 5.0 
         & 19975 & 16.6 \\      

         \midrule
         ImageNet-1k~\cite{deng2009imagenet}
         & \reftab{cls_comp}
         & N/A & 1000
         & 6361 & 6.4 
         & 19741 & 19.7 \\      

         \bottomrule
    \end{tabular}
    \lesspace
    \lbltab{statistics_generated_hiers}
\end{table*}

Being a hierarchy-based method, validating \shine's effectiveness with various hierarchy sources is crucial. In real-world applications, an ideal semantic hierarchy for the target data might not always be available. Therefore, our study focuses on evaluating \shine using not only the dataset-specific class taxonomies~\cite{fan2020few,cole2022label,deng2009imagenet} (the ground-truth hierarchies provided by the datasets as described in \refsec{supp_data}) but also hierarchies synthesized for the target test vocabulary via \llmlong{s} (\llm{s}).
Our key idea is that encyclopedic textual information about semantic categories is readily available on the Internet. Contemporary \llm{s} like ChatGPT~\cite{chatgpt}, trained on vast internet-scale corpora, inherently encode the necessary semantic class taxonomic information in their weights. Similar to the approach used in \chils~\cite{novack2023chils}, we employ an \llm to automatically generate simple three-level semantic hierarchies for the target vocabularies.
We use the ChatGPT~\cite{chatgpt} gpt-3.5-turbo model as our \llm via its public API to generate the synthetic semantic hierarchy with a temperature parameter of 0.7.
{See our \href{https://github.com/naver/shine}{codebase} for the generated hierarchies}.

In the following subsections, we first detail the process of prompting \llm{s} to generate hierarchies (\refsec{supp_hier_prompt}) and then summarize the statistics of the hierarchies we generated (\refsec{supp_hier_stat}).

\subsection{Prompting LLMs}
\lblsec{supp_hier_prompt}
In scenarios where a ground-truth hierarchy is unavailable, and given a label vocabulary $\classTest$ representing the target Classes of Interest (CoIs) at a specific granularity level of the evaluation dataset, the true super-/sub-categories for each CoI are \textit{unknown}. To generate a simple 3-level hierarchy for $\classTest$, we first use ChatGPT~\cite{chatgpt} to generate a list of super-categories for each CoI $c \in \classTest$ using the following \textbf{super-category prompt}:

\begin{center}
    \mytexttt{Generate a list of \textbf{p} super-categories that the following {\textbf{[context]}} object belongs to and output the list separated by '\&': $\mathbf{c}$}
\end{center}

where $p=3$. Subsequently, following Novack \etal~\cite{novack2023chils}, for each CoI $c \in \classTest$, we query ChatGPT~\cite{chatgpt} 
to generate a list of sub-categories using the
following \textbf{sub-category prompt}:

\begin{center}
    \mytexttt{Generate a list of \textbf{q} types of the following {\textbf{[context]}} and output the list separated by '\&': $\mathbf{c}$}
\end{center}

where $q=10$. The \mytexttt{\textbf{\textbf{[context]}}} prompt is consistently set to \mytexttt{\textbf{object}} across all datasets, except for \inat~\cite{cole2022label}, where context-specific prompts like \mytexttt{\textbf{species}} or \mytexttt{\textbf{genus}} are used, aligning with its biological tree of life structure. The \mytexttt{'\&'} symbol serves as a separator prompt, facilitating the formatting of ChatGPT's responses for easier post-parsing of category names. Moreover, the final lists of super-categories and sub-categories are the \textit{union} of results from $t=3$ \llm{} queries. To be more specific, we employ the same super-/sub-category prompts for querying the \llm{} $t=3$ times for each target CoI, and then amalgamate these \llm{} responses to form the final results.

In order to generate hierarchies for all datasets, we fix $p=3$, $q=10$, and $t=3$. It is important to note that we did not perform any extensive hyperparameter tuning for $p$, $q$, and $t$, as our goal is to construct hierarchies automatically and validate \shine's effectiveness with open and noisy hierarchies. Apart from parsing category names from ChatGPT's responses, we do not perform any additional cleaning or organizing of the query results, ensuring an unbiased evaluation of our method's inherent efficacy. The hierarchies generated for the evaluation datasets are directly employed as \llm-generated hierarchies by \shine in our experiments to assess its performance.

\myparagraph{Discussion: Differences between our hierarchy generation process and the one from \chils~\cite{novack2023chils}.}
For any given target vocabulary, \chils uses GPT-3 to generate only sub-categories, forming a two-level hierarchy. In our work, we adopt the \textbf{sub-category prompt} from \chils for generating sub-categories. However, our hierarchy generation strategy significantly differs from \chils in three key respects: \textit{i)} We generate both super-categories and sub-categories, creating a more comprehensive three-level hierarchy. \textit{ii)} We query our \llm three times ($t=3$) and use the union of the outcomes of these queries as the final set, aiming to enrich and diversify the category sets with varied categorization principles. \textit{iii)} As a result of merging and de-duplicating the generated category names from three \llm{} queries, we do not have a predetermined (fixed) number of super-/sub-categories for each target CoI (class). Thus, our generated hierarchies are more varied and imbalanced, aligning more closely with real-world scenarios.

\myparagraph{Discussion: The rationale behind generating $p=3$ super-categories instead of just one.}
In real-world contexts, there is no single ``optimal" hierarchy for any given vocabulary set. A single vocabulary can have multiple, equally valid hierarchical arrangements, depending on the categorization principles applied. For example, \mytexttt{"Vegetable salad"} might be classified under various super-categories—such as \mytexttt{"Appetizer"}, \mytexttt{"Cold dish"}, \mytexttt{"Side dish"}, or simply \mytexttt{"Vegetable"}—based on cultural or contextual differences. Therefore, \textbf{a truly robust and effective hierarchy-based method should function with hierarchies open to diverse categorization principles.} In such open hierarchies, categories are open to multiple categorization principles (\ie, one class may link to several super-category nodes). Thus, we choose to generate $p=3$ super-categories per target CoI (category) in $\classTest$ at each single \llm{} query.
In our 3-level synthetic hierarchies, each target CoI falls under multiple super-categories generated from three times of \llm{} queries, reflecting various and diverse categorization principles. This approach allows us to rigorously evaluate the efficacy of our proposed \shine in realistic, diverse yet noisy categorization scenarios.

\subsection{Summary statistics of the \llm hierarchies}
\lblsec{supp_hier_stat}
In \reftab{statistics_generated_hiers}, we present comprehensive summary statistics for the synthetic hierarchies generated by the \llm{} across each dataset at every label vocabulary level. All synthetic hierarchies are created using $p=3$ and $q=10$, with the final super-/sub-categories a de-duplicated union of results from $t=3$ \llm{} queries. As shown in \reftab{statistics_generated_hiers}, the hierarchies synthesized are both highly open (each CoI is linked to multiple super-categories) and noisy (sub-categories might not be present in the dataset). Despite these challenges, as shown in \reftab{imprv_detic_swin}, \reftab{imprv_vldet_codet_swinb}, \reftab{cls_comp}, \reftab{base_novel_comp}, and \reffig{expanded_comp}, \shine performs effectively using such open and noisy synthetic hierarchies, consistently improving the baseline results. This underlines the adaptability and robustness of \shine in using open and noisy semantic hierarchies when the ground-truth hierarchies are not available.

\section{Further Implementation Details of \shine}
\lblsec{supp_impl}

\subsection{Hierarchy-aware Sentences Integration}
\lblsec{supp_impl_toyexp}

\begin{figure*}[!t]
    \centering
    \includegraphics[width=\linewidth]{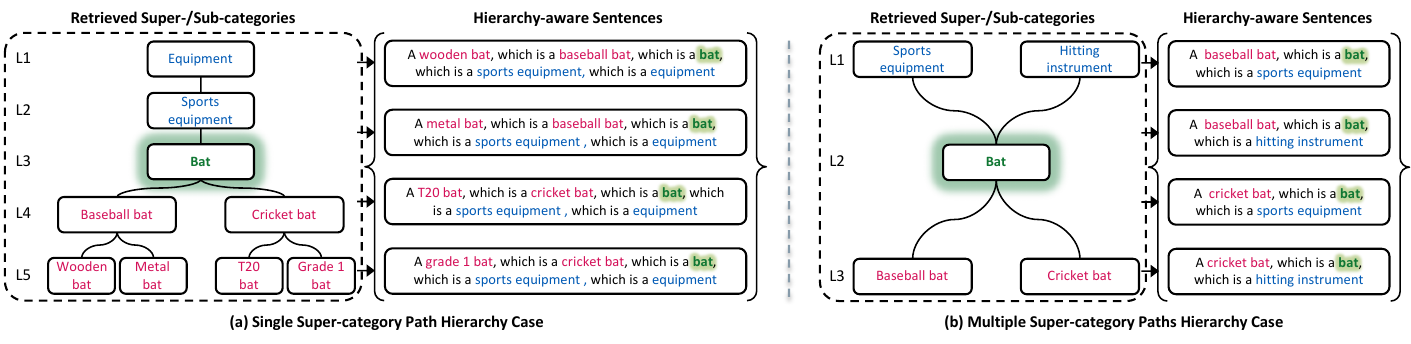}
    \lesspace
    \caption{
    Examples of integrating hierarchy-aware sentences with different hierarchy structures. We use \mytexttt{{"\color{me_node}{\textbf{Bat}}}"} as an example of the target Class of Interest (CoI) for example. The retrieved super-/sub-categories and the target CoI are color-coded in {\color{sup_node} blue} and {\color{sub_node} red}, and {\color{me_node} green}, respectively. \textbf{(a)} The target CoI is linked to a unique super-category at each higher hierarchy level and multiple sub-categories at each lower level, akin to the ground-truth hierarchy structure of the datasets. \textbf{(b)} The target CoI is associated with multiple super-categories at the upper hierarchy level and multiple sub-categories at the lower level, akin to the simple three-level \llm-generated hierarchy structures.
    }
    \lesspace
    \lblfig{explain_hier}
\end{figure*}

This section provides a further explanation of \shine's process for integrating hierarchy-aware sentences with different hierarchical structures, as illustrated in \reffig{explain_hier}.

\myparagraph{Single super-category path hierarchy case (ground-truth hierarchy structures).}
\reffig{explain_hier}(a) illustrates the case where the target Class of Interests (CoI) is linked to a unique super-category at each higher hierarchical level and multiple sub-categories at each lower level. In this case, \shine employs the \textbf{Is-A} connector to form hierarchy-aware sentences by integrating the lowest linked sub-category, the target CoI, and the highest super-category, following their hierarchical relationships in a bottom-up manner. As a result, the total number of constructed sentences in this case equals the number of the lowest linked sub-categories.

\myparagraph{Multiple super-category path hierarchy case (LLM-generated hierarchy structures).} \reffig{explain_hier}(b) displays the case where the target CoI is linked to multiple super-categories at the upper level and several sub-categories at the lower level. Here, \shine builds hierarchy-aware sentences by iterating through all combinations of the linked sub-categories, super-categories, and the target CoI. The \textbf{Is-A} connector is used to connect these categories in a specific-to-abstract order. The resulting number of constructed sentences in this case equals the product of the counts of the lowest linked sub-categories and the linked super-categories.

\subsection{Pseudo-code of \shine}
\lblsec{supp_impl_code}

We show the pseudocode for the core implementation of \shine in \refalg{code_appraoch}, tailored for a three-level hierarchy.

\begin{algorithm}[!ht]
\caption{Pseudocode for constructing \shine classifier \textit{offline} for \ovod detectors in a PyTorch-like style.}
\lblalg{code_appraoch}

\definecolor{codeblue}{rgb}{0.25,0.5,0.8}
\lstset{
  backgroundcolor=\color{white},
  basicstyle=\fontsize{7.2pt}{7.2pt}\ttfamily\selectfont,
  columns=fullflexible,
  breaklines=true,
  captionpos=b,
  commentstyle=\fontsize{7.2pt}{7.2pt}\color{codeblue},
  keywordstyle=\fontsize{7.2pt}{7.2pt},
}
\begin{lstlisting}[language=python]
# target_vocabulary: input class vocabulary for
# inference
# shine_classifier: output SHiNe classifier for
# OvOD detectors
# hrchy: a semantic hierarchy for the target
# vocabulary
# aggregator: computes mean vector or principal
# eigenvector of the given embeddings
# tokenizer: tokenizes given text
# text_encoder: VLM text encoder

# container for SHiNe
shine_classifier = []

# the proposed Is-A connector
isa_connector = "which is a"

# build SHiNe classifier weight vector for each
# class in the vocabulary
for class_name in target_vocabulary:
    # retrieve super-category names
    super_names = hrchy.get_parents(class_name)
    # retrieve sub-category names
    sub_names = hrchy.get_children(class_name)

    # form specific-to-abstract branches combining
    # super-/sub-categories, and the target class
    # name
    branches = [
        [sub_name, class_name, super_name]
        for super_name in super_names
        for sub_name   in child_names
    ]

    # construct hierarch-aware sentences in natural
    # language using the Is-A connector
    sentences = [
        f"a {branch[0]}"
        + "".join([
        f", {isa_connector} {name}"
        for name in branch[1:]
        ])
        for branch in branches
    ]

    # tokenize the sentences
    text_tokens = tokenizer(sentences)
    # extract textual feature representations
    text_embeddings = text_encoder(text_tokens)
    
    # fuse the embeddings into a single nexus-based
    # classifier vector
    nexus_vector = aggregator(text_embeddings)
    
    # append the single classifier vector to the 
    # classifier container
    shine_classifier.extend(nexus_vector)

# stack all the constructed classifier vectors as
# the SHiNe classifier
shine_classifier = torch.stack(shine_classifier)

# l2-normalize the classifier vectors
shine_classifier = l2_normalize(shine_classifier,
                                           dim=1)

# the shine_classifier is output and applied
# directly to the OvOD detector
\end{lstlisting}
\end{algorithm}

\subsection{Time Complexity Analysis of \shine}
\lblsec{supp_impl_time}
Let $c$ be the number of Classes of Interest (CoIs) in a given vocabulary, and let $p$ and $q$ represent the average number of related super-categories and sub-categories per CoI, respectively, in a hierarchy. Our proposed method, \shine, aggregates hierarchy-aware information from both super-categories and sub-categories into $c$ \textit{nexus}-based embeddings (offline).
Consequently, at inference, both memory and time complexity of \shine scale linearly as $\mathcal{O}(c)$. It is important to note that this scalability at inference is unaffected by the number of related super-/sub-categories, because they are only used offline to generate $\bnexus_{c}$. The offline pipeline to construct \shine \ovod classifier needs to run only once.

In contrast, the time and memory complexities for \chils~\cite{novack2023chils} scale at inference as $\mathcal{O}(c(1 + q))$, because image-text similarity scores are computed for vocabulary nodes \textit{and} all their children. \hclip~\cite{ge2023improving}, on the other hand, involves a search procedure \textit{online} across $p \cdot (c+1)$ prompt combinations for the top $k$ (\eg, $k=5$) predicted CoIs, resulting in a time complexity of $\mathcal{O}(c + p \cdot (q+1) \cdot k)$. Crucially, the operations for $p \cdot (q+1) \cdot k$ only commence after the prediction based on the first $c$ standard prompts. Unlike \shine and \chils~\cite{novack2023chils}, for which the embeddings are precomputed and the class predictions can be fully parallelized, \hclip requires encoding the latter $p \cdot (q+1) \cdot k$ CLIP~\cite{radford2021learning} text embeddings at test time on-the-fly. Furthermore, it employs a search-on-the-fly mechanism, resulting in significant computational overheads. This makes \hclip a sub-optimal candidate for many applications, particularly those like detection and segmentation tasks that require per-box, per-mask, or even per-pixel prediction.

Given the extensive number of super-/sub-categories in the hierarchy employed in our experiments, as detailed in \reftab{statistics_generated_hiers}, the substantial computational overheads imposed by \chils and \hclip become evident.

\subsection{Implementation Details of Aggregators}
\lblsec{supp_impl_peigen}

\myparagraph{Mean-aggregator.}
During the semantic hierarchy \textit{nexus} classifier construction phase, as illustrated in \reffig{approach}(3), \shine, by default, uses Eq.~\refeq{eq:aggregation} where the ``Aggregator'' is the mean operation, as
\begin{align}
    \bnexus_c &=
    \frac{1}{K}
    \sum_{k=1}^{K}
    \enctxt\left(\sent_k^c\right)\enspace ,
    \lbleq{mean_aggr}
\end{align}
where $\enctxt$ is the frozen CLIP~\cite{radford2021learning} text encoder, and $\{\sent_k^c\}_{k=1}^{K}$ represents the $K$ hierarchy-aware sentences, which are built by integrating all super-/sub-categories related to the target class (CoI) $c$ using our proposed \textbf{Is-A} connector. This aggregator, which we call the \textbf{mean-aggregator}, calculates the mean of the encoded sentences' embeddings to form the final \textit{nexus}-based classifier weight vector for $c$. This mean vector is the centroid represented within CLIP's embedding space, summarizing the general characteristics of the hierarchy-aware embeddings related to the target CoI.
At inference, the classification decision for a region is based on the cosine similarity between the visual embedding of the region and the hierarchy-aware representation defined by the mean vector $\bnexus$, which we call \textit{nexus}. This approach renders the decision-making process less sensitive to variations in the semantic granularity of the name $c$. Note that all the embeddings are $l$2-normalized.

\myparagraph{Principal Eigenvector Aggregator}
Drawing inspiration from text classification techniques in Natural Language Processing (NLP)~\cite{li1998classification, shin2018interpreting, gewers2021principal}, we introduce an alternative aggregation approach, called the \textbf{principal eigenvector aggregator}. This method uses the principal eigenvector of the sentence embeddings matrix as the classifier weight vector $\bnexus_c$. Specifically, for a set of hierarchy-aware sentences $\{\sent_{k}^{c}\}_{k=1}^{K}$, we first apply a Singular Vector Decomposition (SVD) operation on their embedding matrix as:
\begin{align}
    \rmU \rmS \rmV^{T} = 
    \text{SVD}\left(
        \textrm{concat}_{k=1}^{K} \left\{ \enctxt\left(\sent_k^c\right)
        \right\}
    \right) \enspace ,
    \lbleq{peigen_aggr}
\end{align}
where $\rmU$ and $\rmV$ are orthogonal matrices representing the left and right singular vectors, respectively, and $\rmS$ is a diagonal matrix with singular values in descending order. Subsequently, we can derive the principal eigenvector, corresponding to the largest singular value in the sentence embedding matrix, by selecting the first column of matrix $\rmV$ as:
\begin{align}
    \bnexus_c=\rmV[:, 0] \enspace ,
\end{align}
where $\bnexus_c$ serves as the \textit{nexus}-based classifier vector for the target class $c$. In contrast to the mean-aggregator, the \textbf{principal eigenvector aggregator} captures the dominant trend in the sentence embeddings (as known as their ``theme'', to maintain NLP terminology), to effectively represent the CoIs. Note that all the embeddings are $l$2-normalized.

Next, we explain the rationale behind this aggregator design.
In high-dimensional semantic spaces like the 512-dimensional vision-language aligned embedding space of CLIP ViT-B/32, the principal eigenvector is able to capture the most significant semantic patterns or trends within the embeddings. This approach stems from the understanding that the direction of greatest variance in the space contains the most informative representation of semantic embeddings. Projecting the high-dimensional hierarchy-aware sentence embeddings of a target class (CoI) onto this principal eigenvector yields a condensed yet information-rich representation, preserving the essence of the original hierarchy-aware sentences. Consequently, during inference, classification decisions for a region are based on the cosine similarity between the region's embedding and the semantic pattern or trend depicted by the principal eigenvector. This differs from the representation centroid approach used by the mean-aggregator.

We compare the \textbf{mean-aggregator} and the \textbf{principal eigenvector aggregator} in \refsec{expt_selfstudy} of the main paper. While the principal eigenvector aggregator shows slightly lower performance compared to the mean-aggregator in general, its potential application in \vlm tasks might be interesting for future research. In general, given the intimate connection between computer vision and NLP in open-vocabulary models, we believe in the importance of enabling more connections between the two fields---in this case, drawing from the NLP field of topic modeling.

\section{Extended Analysis of \shine on FSOD}
\lblsec{supp_expt_ablation_fsod}

\begin{figure*}[!t]
    \centering
    \includegraphics[width=\linewidth]{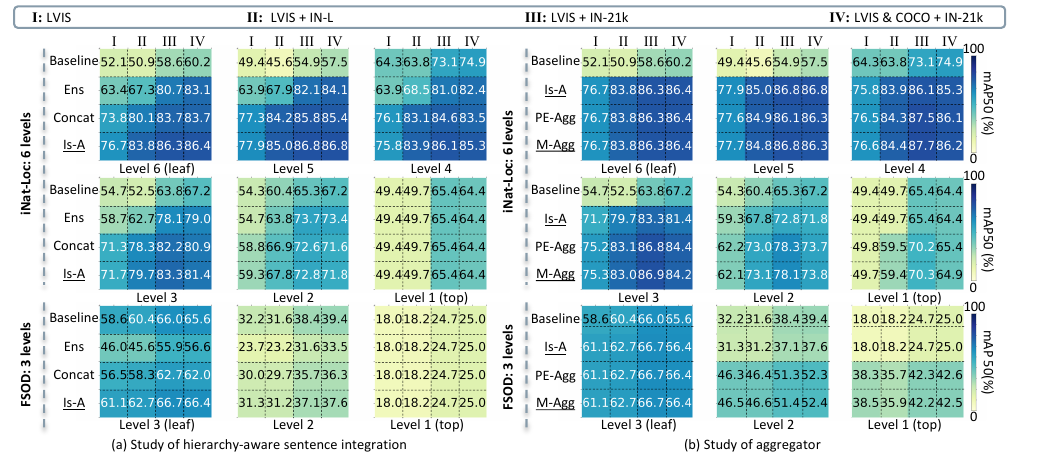}
    \vspace{-7mm}
    \caption{
    Further study of hierarchy-aware sentence integration methods \textbf{(left)} and aggregators \textbf{(right)} across various label granularity levels on both \inat and \fsod datasets. Darker color indicates higher mAP50. Components used by default in \shine are underlined. Detic~\cite{zhou2022detecting} with Swin-B backbone, trained using various combinations of supervisory signals described in \reftab{training_data}, serves as the baseline open-vocabulary detector for all methods evaluated. To evaluate the effectiveness of hierarchy-based components, we use the ground-truth hierarchy for all methods that rely on hierarchies.
    }
    \lblfig{self_study_both}
\end{figure*}

In \reffig{self_study_both}, we present an expanded study of the core components of \shine, examining their effectiveness across various levels of label granularity on both the \inat and \fsod datasets. The results from \fsod align with those observed in the \inat-only study shown in \reffig{self_study} of the main paper. Next, we provide further analysis of \shine's core components. 

\myparagraph{Extended discussion: the Is-A connector effectively integrates hierarchy knowledge in natural sentences.}
The effectiveness of the proposed \textbf{Is-A} connector is studied in \reffig{self_study_both}(a). Excluding the top (abstract) levels where all methods, including \textbf{Ens}, \textbf{Concat}, and \textbf{Is-A}, revert to the plain baseline due to the absence of further parent nodes, the methods leveraging super-category information consistently outperform the baseline across nearly all levels of granularity. This improvement is attributed to directing the model’s focus towards more general concepts via super-category-inclusive classifiers. An exception occurs at the second level of \fsod (\reffig{self_study_both}(a-FSOD-L2)), where no method exceeds the baseline. We speculate that at this level, target categories like \mytexttt{"Fruit"} are already highly abstract, rendering the addition of more abstract parent categories like \mytexttt{"Food"} redundant in clarifying ambiguities. Nevertheless, this challenge is alleviated when sub-categories are also included in the aggregation step. In comparative terms, the \textbf{Is-A} and \textbf{Concat} connectors yield greater gains than \textbf{Ens}, highlighting the advantage of capturing internal semantic relationships for distinguishing between classes. Notably, our \textbf{Is-A} connector surpasses \textbf{Concat} at all levels of granularity in both datasets, improving the baseline mAP50 by up to \textbf{\color{higher}+39.4} points on \inat (\reffig{self_study_both}(a-iNat-L5)) and \textbf{\color{higher}+2.5} points on \fsod (\reffig{self_study_both}(a-FSOD-L3)). This indicates the superior effectiveness of \textbf{Is-A}'s explicit modeling of category relationships compared to the mere sequential ordering of class names from specific to abstract by \textbf{Concat}. Overall, the integration of more abstract concepts proves beneficial in object detection across diverse label granularities, with our \textbf{Is-A} connector particularly excelling due to its effective incorporation of hierarchical knowledge into natural language sentences, achieved by explicitly modeling internal category relationships.

\myparagraph{Extended discussion: A simple mean-aggregator is sufficient for hierarchy-aware sentences fusion.}
The impact of the aggregation step is analyzed in \reffig{self_study_both}(b), focusing on both the mean-aggregator (M-Agg) and the principal eigenvector aggregator (PE-Agg). These aggregators consistently outperform the baseline across various models and levels of label granularity in all datasets. Notably, their advantage becomes more pronounced with increasingly abstract target vocabularies, surpassing the benchmarks set by the \textbf{Is-A} method. This is especially evident in cases involving highly abstract label vocabularies, where these aggregation methods significantly improve baseline performance, achieving gains of up to \textbf{{\color{higher}+9.8}} points in \inat (\reffig{self_study_both}(b-iNat-L1)) and \textbf{{\color{higher}+20.5}} points in \fsod (\reffig{self_study_both}(b-FSOD-L1)).

These results underscore the effectiveness of the aggregation step in fusing hierarchy-aware sentences into semantic \textit{nexus}-based classifiers. This fusion allows the \textit{nexus}-based classifier to use both specific knowledge from sub-categories and abstract knowledge from super-categories, thereby improving the baseline detector's ability to discriminate visual object robustly.

\begin{figure*}[!ttt]
    \centering
    \includegraphics[width=0.9\linewidth]{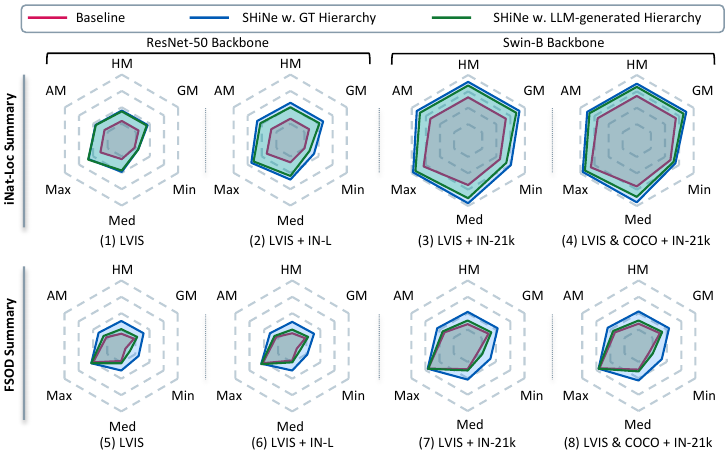}
    \caption{
    Additional summary statistics across all levels for the main experimental results in \reftab{imprv_detic_swin} for \inat \textbf{(upper)} and \fsod \textbf{(lower)}, respectively. This summary includes various measures for mAP50, such as arithmetic mean (AM), harmonic mean (HM), geometric mean (GM), minimum value (Min), median (Med), and maximum value (Max), calculated across all granularity levels within each dataset. A larger area indicates better performance across various metrics. Gray dashed gridlines are scaled from 10 (innermost) to 100 (outermost).
    }
\lblfig{imprv_summary}
\end{figure*}

\section{Comparison with Additional Baselines}
\lblsec{supp_expt_ablation_root_ems}

To further validate the effectiveness of the proposed Is-A prompting method, we further compare \shine with two additional baselines: \textit{i)} \textbf{Root-Stmt} prompting, which explicitly states the root (target) class and its super/sub-classes using the template like \mytexttt{"A {\color{me_node}bat}, which is a {\color{sup_node}sports equipment} and can be instantiated in a {\color{sub_node}wooden baseball bat} or a {\color{sub_node}baseball bat}"}; \textit{ii)} \textbf{80-Prompts}, where we embed the target class name into the 80 hand-crafted prompts from CLIP~\cite{radford2021learning} and average the scores. As shown in \reftab{comp_additional_baselines}, methods leveraging a hierarchy consistently 
surpass the 80-prompt ensemble baseline, demonstrating the benefits of leveraging hierarchy knowledge. 
Moreover, \shine's superior performance to the Root-Stmt baseline suggests that 
Is-A prompting and nexus aggregation is more effective for combining hierarchy information.
\begin{table}[hhh!]
    \centering
    \small
    \tablestyle{0.7pt}{0.6}
    \caption{
    Comparison with additional baseline methods on \inat and \fsod datasets. Detic with Swin-B backbone trained with LVIS and IN-21k is used as baseline. mAP50 is reported. 
    }
    \lbltab{comp_additional_baselines}
    \begin{tabular}{cl|ccccccr|cccr|}
         \toprule
         
         \multirow{2}{*}{\rotatebox[origin=c]{90}{Hrchy}}
         & \multicolumn{1}{l}{Set}
         & \multicolumn{7}{c}{\inat@mAP50} 
         & \multicolumn{4}{c}{\fsod@mAP50} \\

         \cmidrule(r){2-2}
         \cmidrule(r){3-9}
         \cmidrule(r){10-13}
         
         & Level
         & L6 & L5 & L4 & L3 & L2 & L1 
         & \multicolumn{1}{c|}{$\overline{\Delta}$}
         & L3 & L2 & L1 
        & \multicolumn{1}{c|}{$\overline{\Delta}$}\\

         \cmidrule(r){1-2}
         \cmidrule(r){3-9}
         \cmidrule(r){10-13}
         
         \multirow{2}{*}{\rotatebox[origin=c]{90}{N/A}}
         & Baseline
         & 58.6 & 54.9 & 73.1 & 63.8 & 65.3 & 65.4 
         & \multicolumn{1}{c|}{-}
         & 66.0 & 38.4 & 24.7 
         & \multicolumn{1}{c|}{-} \\

         & 80-Prompts
         & 59.3 & 55.9 & 73.4 & 66.9 & 66.4 & 65.6
         & {\color{higher}+1.1}
         & 66.1 & 38.7 & 26.0 
         & {\color{higher}+0.6} \\
         
         \cmidrule(r){1-2}
         \cmidrule(r){3-9}
         \cmidrule(r){10-13}
         
         \multirow{2}{*}{\rotatebox[origin=c]{90}{GT}} & Root-Stmt
         & 86.3 & 83.1 & 83.9 & 82.6 & 72.1 & 66.6
         & {\color{higher}+15.6}
         & 66.7 & 46.7 & 31.6 
         & {\color{higher}+5.3} \\
         
         & \shine
         & \bf86.3 & \bf86.8 & \bf87.7 & \bf86.9 & \bf78.1 & \bf70.3
         & \bf{\color{higher}+19.2}
         & \bf66.7 & \bf51.4 & \bf42.2 
         & \bf{\color{higher}+10.4} \\
    \bottomrule
    \end{tabular}
    \lesspace
\end{table}

\section{Extended Main Experimental Results}
\lblsec{supp_expt_more_swinb}
\begin{table}[!ht]
    \centering
    \small
    \tablestyle{2.1pt}{1}
    \caption{Additional results are provided for Detic~\cite{zhou2022detecting} with a Swin-B backbone, trained using both \textbf{I}-LVIS and \textbf{II}-LVIS+IN-L supervisory signal combinations. Detection performance across varying label granularity levels on \inat \textbf{(upper)} and \fsod \textbf{(lower)} datasets are reported. \shine is directly applied to the baseline detector (BL)~\cite{zhou2022detecting} with ground-truth (GT-\hierarchy) and LLM-generated (LLM-\hierarchy) hierarchies. mAP50 (\%) is reported.
    }
    
    \begin{tabular}{ll|lll|lll|}
         \toprule
         &\multicolumn{1}{c}{}
         & \multicolumn{6}{c}{Swin-B Backbone} \\
         
         \cmidrule(r){3-8}
         
         &\multicolumn{1}{c}{}
         & \multicolumn{3}{c}{\textbf{I} - LVIS} 
         & \multicolumn{3}{c}{\textbf{II} - LVIS + IN-L} \\
         
         \cmidrule(r){3-5}
         \cmidrule(r){6-8}

         \multirow{2}{*}{\rotatebox[origin=c]{90}{Set}} & \multirow{2}{*}{\rotatebox[origin=c]{90}{Level}} 
         & \multicolumn{1}{c}{\multirow{2}{*}{BL}} & \multicolumn{1}{c}{\shine} & \multicolumn{1}{c|}{\shine} 
         & \multicolumn{1}{c}{\multirow{2}{*}{BL}} & \multicolumn{1}{c}{\shine} & \multicolumn{1}{c|}{\shine} \\
         
         &
         && \multicolumn{1}{c}{(GT-\hierarchy)} & \multicolumn{1}{c|}{(LLM-\hierarchy)} 
         && \multicolumn{1}{c}{(GT-\hierarchy)} & \multicolumn{1}{c|}{(LLM-\hierarchy)} \\

         \cmidrule(r){3-5}
         \cmidrule(r){6-8}
         
         \multirow{6}{*}{\rotatebox[origin=c]{90}{\inat}}

         & L6
         & 52.1 & \bf76.7({\color{higher}+24.6}) & 74.0({\color{higher}+21.9}) 
         & 50.9 & \bf83.8({\color{higher}+32.9}) & 78.8({\color{higher}+27.9}) \\
								
         & L5 
         & 49.4 & \bf77.7({\color{higher}+28.3}) & 68.4({\color{higher}+19.0}) 
         & 45.6 & \bf84.8({\color{higher}+39.2}) & 69.0({\color{higher}+23.4}) \\
 										
         & L4 
         & 64.3 & \bf76.6({\color{higher}+12.3}) & 73.7({\color{higher}+9.4})
         & 63.8 & \bf84.4({\color{higher}+20.6})    & 79.3({\color{higher}+15.5}) \\
										
         & L3 
         & 54.7 & \bf75.3({\color{higher}+20.6}) & 73.2({\color{higher}+18.5})
         & 52.5 & \bf83.0({\color{higher}+30.5}) & 78.7({\color{higher}+26.2}) \\
				
         & L2 
         & 54.3 & 62.1({\color{higher}+7.8}) & \bf62.8({\color{higher}+8.5}) 
         & 60.4 & 73.1({\color{higher}+12.7}) & \bf75.1({\color{higher}+14.7}) \\
 											
         & L1
         & 49.4 & 49.7({\color{higher}+0.3}) & \bf50.7({\color{higher}+1.3}) 
         & 49.7 & \bf59.4({\color{higher}+9.7}) & 49.8({\color{higher}+0.1}) \\
         
         \cmidrule(r){3-5}
         \cmidrule(r){6-8}
         
         \multirow{3}{*}{\rotatebox[origin=c]{90}{\fsod}}
 										
         & L3
         & 58.6 & \bf61.1({\color{higher}+2.5}) & 61.0({\color{higher}+2.4}) 
         & 60.4 & \bf62.7({\color{higher}+2.3}) & 62.1({\color{higher}+1.7}) \\
 									
         & L2 
         & 32.2 & \bf46.5({\color{higher}+14.3}) & 35.6({\color{higher}+3.4}) 
         & 31.6 & \bf46.6({\color{higher}+15.0}) & 33.5({\color{higher}+1.9}) \\
         
         & L1\dag 
         & 18.0 & \bf38.5({\color{higher}+20.5}) & 23.7({\color{higher}+5.7}) 
         & 18.2 & \bf35.9({\color{higher}+17.7}) & 22.3({\color{higher}+4.1}) \\
         
         \bottomrule
    \end{tabular}
    \lesspace
    \lbltab{imprv_detic_swinB_12only}
\end{table}

In \reftab{imprv_detic_swinB_12only}, we present additional experimental results from applying our proposed \shine to Detic~\cite{zhou2022detecting} with a Swin-B~\cite{liu2021swin} backbone, trained using only LVIS and LVIS combined with IN-L~\cite{deng2009imagenet} as auxiliary weak supervisory signals. This observation is consistent with those in \reftab{imprv_detic_swin} from the main paper, demonstrating that \shine consistently and substantially improves the performance of the baseline \ovod detector on both \inat and \fsod datasets. This improvement spans across various label vocabulary granularities and is evident with both the ground-truth hierarchy (GT-H) and a synthetic hierarchy generated by \llm{} (LLM-H).

\section{Summary  of the Main Experiments}
\lblsec{supp_expt_stat}
Beyond the per-level comparison in \reftab{imprv_detic_swin} of the main paper, \reffig{imprv_summary} offers an extended comparison (calculated from \reftab{imprv_detic_swin}) using various summary statistical metrics. As shown in \reffig{imprv_summary}, our proposed \shine consistently and markedly enhances the baseline \ovod detector's performance across a range of summary metrics, including arithmetic mean (AM), harmonic mean (HM), geometric mean(GM), on both datasets. The harmonic and geometric means are employed to present the evaluation results from diverse perspectives, particularly in contexts where extreme values might skew the interpretation. These means are less influenced by extreme values, such as exceptionally high or low mAP50 scores at specific granularity levels. The enhancement from \shine is apparent when employing both the ground-truth hierarchy and a synthetic hierarchy generated by the \llm{}. Notably, \shine most significantly improves the baseline's weakest performance (minimum mAP50), suggesting a notable improvement in performance consistency by improving the minimum achieved performance across granularity levels. These results demonstrate that \shine not only boosts overall performance but also enhances consistency across different vocabulary granularities, a crucial aspect for real-world applications.

\section{Experiments with other OvOD Detectors}
\lblsec{supp_expt_oth_ovod}

\begin{table}[!t]
    \centering
    \small
    \tablestyle{1.9pt}{0.8}
    \caption{
    Comparison with CORA~\cite{wu2023cora} and VLDet (\vldet)~\cite{lin2022learning} 
    on \inat and \fsod.
    \shine is applied to the baseline methods, respectively.
    All methods employ ResNet-50~\cite{he2016deep} as backbone. Note that CORA uses only box-annotated COCO~\cite{lin2014microsoft} base split for training, while VLDet uses box-annotated LVIS~\cite{gupta2019lvis} and image-caption-annotated CC3M~\cite{sharma2018conceptual} as supervisory signals. mAP50 (\%) is reported.
    }
    \begin{tabular}{ll|lll|lll|}
         \toprule

         \multirow{2}{*}{\rotatebox[origin=c]{90}{Set}} 
         & \multirow{2}{*}{\rotatebox[origin=c]{90}{Level}} 
         & \multicolumn{1}{c}{\multirow{2}{*}{CORA}} & \multicolumn{1}{c}{\shine} & \multicolumn{1}{c|}{\shine} 
         & \multicolumn{1}{c}{\multirow{2}{*}{\vldet}} & \multicolumn{1}{c}{\shine} & \multicolumn{1}{c|}{\shine} \\
         
         &
         && \multicolumn{1}{c}{(GT-\hierarchy)} & \multicolumn{1}{c|}{(LLM-\hierarchy)}
         &&\multicolumn{1}{c}{(GT-\hierarchy)} & \multicolumn{1}{c|}{(LLM-\hierarchy)}\\

         \cmidrule(r){3-5}
         \cmidrule(r){6-8}
         
         \multirow{6}{*}{\rotatebox[origin=c]{90}{\inat}}

         & L6
         & 31.2 & 54.2({\color{higher}+23.0}) & \bf54.8({\color{higher}+23.6}) 
         & 48.9 & 62.4({\color{higher}+13.5}) & \bf64.8({\color{higher}+15.9})\\ 
								
         & L5
         & 22.6 & \bf51.9({\color{higher}+29.3}) & 35.7({\color{higher}+13.1}) 
         & 44.3 & \bf60.6({\color{higher}+16.3}) & 52.6({\color{higher}+8.3})\\ 
 										
         & L4
         & 21.7 & \bf50.7({\color{higher}+29.0}) & 36.2({\color{higher}+14.5}) 
         & 42.9 & \bf58.7({\color{higher}+15.8}) & 53.6\bf({\color{higher}+10.7})\\  

         & L3
         & 26.0 & \bf50.5({\color{higher}+24.5}) & 43.4({\color{higher}+17.4})
         & 43.8 & \bf63.5({\color{higher}+19.7}) & 58.7({\color{higher}+14.9})\\ 
				
         & L2 
         & 20.0 & \bf33.2({\color{higher}+13.2}) & 24.8({\color{higher}+4.8}) 
         & 34.3 & \bf54.4({\color{higher}+20.1}) & 46.9\bf({\color{higher}+12.6})\\  
 											
         & L1
         & 18.3 & 16.2({\color{lower}-2.1}) & 13.0({\color{lower}-5.3}) 
         & 37.0 & \bf43.2({\color{higher}+6.2}) & 38.6({\color{higher}+1.6})\\ 
         
         \cmidrule(r){3-5}
         \cmidrule(r){6-8}
         
         \multirow{3}{*}{\rotatebox[origin=c]{90}{\fsod}}
         
         & L3
         & 49.3 & \bf51.4({\color{higher}+2.1}) & 51.1({\color{higher}+1.8}) 
         & 48.8 & \bf53.8({\color{higher}+5.0}) & 53.6({\color{higher}+4.8})\\ 
 									
         & L2
         & 21.9 & \bf33.6({\color{higher}+11.7}) & 23.5({\color{higher}+1.6}) 
         & 23.5 & \bf39.3({\color{higher}+15.8}) & 29.8({\color{higher}+6.3})\\ 
         
         & L1
         & 11.6 & \bf26.2({\color{higher}+14.6}) & 14.1({\color{higher}+2.5})
         & 12.8 & \bf31.0({\color{higher}+18.2}) & 17.3({\color{higher}+4.5})\\  
         
         \bottomrule
    \end{tabular}
    \lesspace
    \lbltab{imprv_vldet_cora_rn50}
\end{table}

We further assess \shine{'s performance on top of} 
an additional OvOD detector, CORA~\cite{wu2023cora}, and present the results alongside VLDet~\cite{lin2022learning} with ResNet-50~\cite{he2016deep} in \reftab{imprv_vldet_cora_rn50}. These results and improvements are consistent with those using Detic~\cite{zhou2022detecting}, CoDet~\cite{ma2024codet}, and VLDet~\cite{lin2022learning}, further validating \shine's effectiveness.

\section{Further Experiments on COCO/LVIS}
\lblsec{supp_expt_cocolvis}

\begin{table}[!hhh]
    \centering
    \small
    \tablestyle{5.1pt}{1}
    \caption{
    Comparison of detection performance on COCO and LVIS benchmarks using the OVE protocol. We use Detic~\cite{zhou2022detecting} with a ResNet-50 backbone as the baseline detector (BL). \shine is applied to the baseline using hierarchies generated by \llm{}. All models receive strong supervision on the base class partitions of both datasets, with box-class annotations. A comparison of different weak supervisory signals is also included. mAP50$_{novel}$ and mAP$_{novel}$ denote performance evaluated on the novel class partitions (17 classes for COCO and 337 classes for LVIS), while mAP50$_{all}$ and mAP$_{all}$ represent evaluations on both base and novel classes (65 classes for COCO and 1203 classes for LVIS).
    }
    \begin{tabular}{ll|ll|ll|}
         \toprule
         
         \multirow{5}{*}{\rotatebox[origin=c]{90}{COCO}}& \multicolumn{1}{c}{\multirow{2}{*}{$\dataWeak$}} & \multicolumn{2}{c}{mAP50$_{novel}$} & \multicolumn{2}{c}{mAP50$_{all}$} \\
         \cmidrule(r){3-4}
         \cmidrule(r){5-6}
         &  & \multicolumn{1}{c}{BL} & \multicolumn{1}{c|}{\shine} & \multicolumn{1}{c}{BL} & \multicolumn{1}{c|}{\shine} \\
         \cmidrule(r){2-2}
         \cmidrule(r){3-4}
         \cmidrule(r){5-6}
         & N/A
         & 1.3 & \bf3.2({\color{higher}+1.9}) & 39.3 & \bf39.8({\color{higher}+0.5}) \\
         & COCO Captions 
         & 24.0 & \bf24.3({\color{higher}+0.3}) & 44.8 & \bf44.9({\color{higher}+0.1}) \\
         
         \hline
         
         \multirow{6}{*}{\rotatebox[origin=c]{90}{LVIS}}& \multicolumn{1}{c}{\multirow{2}{*}{$\dataWeak$}} & \multicolumn{2}{c}{mAP$_{novel}$} & \multicolumn{2}{c}{mAP$_{all}$} \\
         
         \cmidrule(r){3-4}
         \cmidrule(r){5-6}
         
         &  & \multicolumn{1}{c}{BL} & \multicolumn{1}{c|}{\shine} & \multicolumn{1}{c}{BL} & \multicolumn{1}{c|}{\shine} \\
         \cmidrule(r){2-2}
         \cmidrule(r){3-4}
         \cmidrule(r){5-6}
         & N/A 
         & 17.6 & \bf20.9({\color{higher}+3.3}) & 33.3 & \bf33.6({\color{higher}+0.3}) \\
         & IN-L
         & \bf26.7 & 25.5({\color{lower}-1.2}) & \bf35.8 & 35.3({\color{lower}-0.5}) \\
         & Conceptual Captions 
         & 19.3 & \bf21.5({\color{higher}+2.2}) & 33.4 & \bf33.5({\color{higher}+0.1}) \\
         \bottomrule
    \end{tabular}

    \lbltab{base_novel_comp}
\end{table}

This section extends the evaluation of \shine to COCO~\cite{lin2014microsoft} and LVIS~\cite{gupta2019lvis}, following the open-vocabulary evaluation (OVE) protocol as described in~\cite{zhu2023survey}. According to the OVE protocol, datasets are divided into base and novel classes; models are trained on base classes with bounding box annotations and then evaluated on novel classes and their union. The base classes are disjoint from the novel classes. We follow the base/novel class partitions for COCO and LVIS as used in \cite{zhou2022detecting}. Both datasets have a single, flat class vocabulary: COCO with 65 classes (48 base, 17 novel) and LVIS with 1203 classes (866 base, 337 novel).
\begin{figure*}[!ttt]
    \centering
    \includegraphics[width=0.9\linewidth]{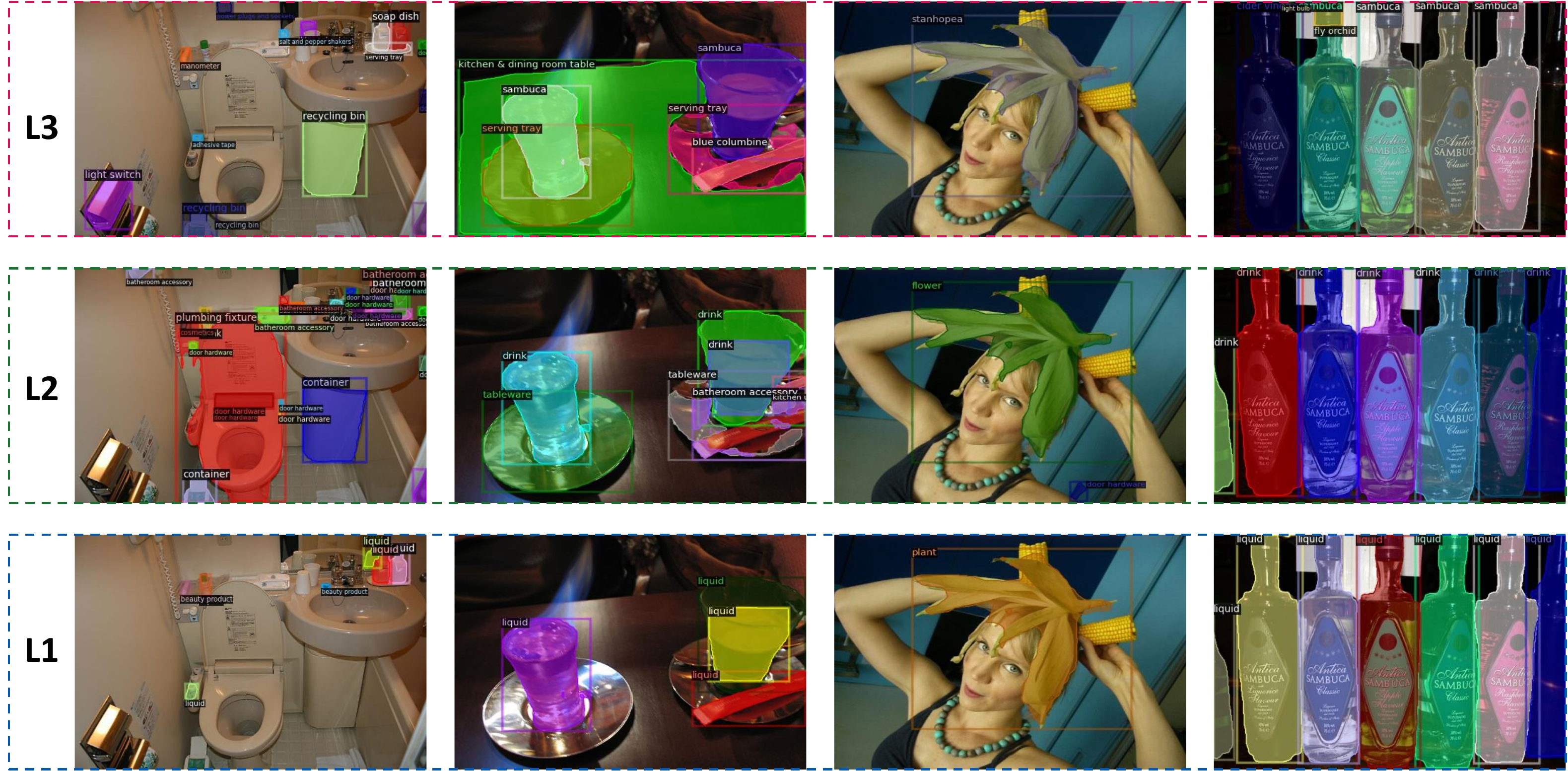}
    \caption{
    Qualitative detection results of \shine applied to Detic~\cite{zhou2022detecting} with Swin-B~\cite{liu2021swin}, evaluated on the \fsod~\cite{fan2020few} dataset across three different label granularity levels. All models are trained using the LVIS + IN-L dataset as strong and weak supervisory signals, respectively. It is advisable to zoom in for a clearer view.
    }
    \lesspace
\lblfig{viz_fsod}
\end{figure*}
We use Detic~\cite{zhou2022detecting} with a ResNet-50~\cite{he2016deep} backbone, trained on the box-class annotated base classes with various weak supervisory signals, as the baseline \ovod detector in this experiment. Specifically, the baseline is trained on COCO-base with 48 classes or LVIS-base with 866 classes. We explore three types of weak supervisory signals as proposed in \cite{zhou2022detecting}: \textit{i)} \textbf{N/A}, using only strong supervisory signals; \textit{ii)} \textbf{IN-L}, a 997-class subset of ImageNet-21k~\cite{deng2009imagenet} intersecting with the LVIS vocabulary; \textit{iii)} Conceptual Captions~\cite{sharma2018conceptual} dataset; and \textit{iv)} COCO Captions~\cite{zhou2022detecting} dataset. For Conceptual Captions and COCO Captions, nouns are parsed from the captions, and both image labels and captions are used for weak supervision~\cite{zhou2022detecting}. We report mAP50 for COCO and the official mask mAP metric for LVIS as suggested in~\cite{gupta2019lvis}.

We evaluate and compare \shine with the baseline under the OVE protocol. In the absence of available ground-truth hierarchy information, we use the \llm{} to generate simple 3-level synthetic hierarchies for the target vocabularies of COCO and LVIS, as described in \reftab{statistics_generated_hiers}. Consequently, \shine is constructed using these generated hierarchies. As shown in the OVE evaluation results in \reftab{base_novel_comp}, \shine notably improves the performance of the baseline detector on both COCO and LVIS benchmarks under the OVE protocol. Interestingly, \shine yields a greater performance gain on the novel class partitions. However, this advantage becomes less pronounced when assessing combined base and novel classes. This is attributed to the model overfitting on the base classes to the text classifier based on the standard \mytexttt{"a \{Class Name\}"} prompts during strongly supervised training. Replacing this overfit classifier with the \shine classifier leads to significant gains on novel class partitions, but slightly reduces performance on base class partition test data. Nevertheless, the consistent improvements achieved by \shine across most cases in \reftab{base_novel_comp} underscore its effectiveness on the COCO and LVIS benchmarks.

\section{Qualitative Analysis of SHiNe}
\lblsec{supp_qua}
In \reffig{viz_fsod} and \reffig{viz_inat}, we showcase the qualitative detection results of \shine when applied to Detic~\cite{zhou2022detecting} across various label granularity levels on the \fsod and \inat datasets. For each granularity level, the same confidence threshold is consistently applied.

\clearpage

\begin{figure*}[!ttt]
    \centering
    \includegraphics[width=0.9\linewidth]{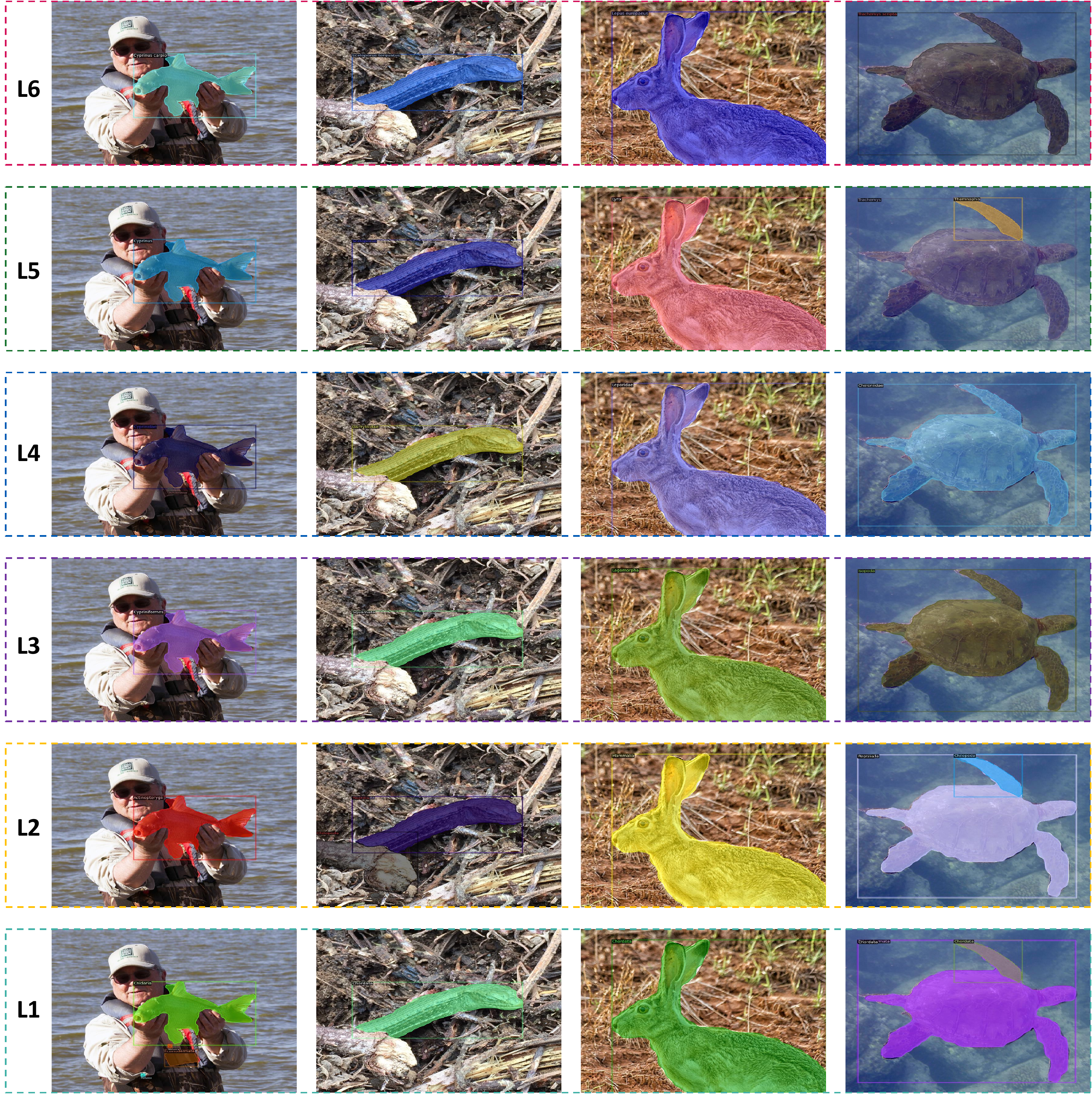}
    \caption{
    Qualitative detection results of \shine applied to Detic~\cite{zhou2022detecting} with Swin-B~\cite{liu2021swin}, evaluated on the \inat~\cite{cole2022label} dataset across six different label granularity levels. All models are trained using the LVIS + IN-L dataset as strong and weak supervisory signals, respectively. It is advisable to zoom in for a clearer view.
    }
\lblfig{viz_inat}
\end{figure*}

\end{document}